\documentclass[lettersize,journal]{IEEEtran}
\usepackage{amsmath,amsfonts}
\usepackage{algorithmic}
\usepackage{algorithm}
\usepackage{array}
\usepackage[caption=false,font=normalsize,labelfont=sf,textfont=sf]{subfig}
\usepackage{textcomp}
\usepackage{stfloats}
\usepackage{url}
\usepackage{verbatim}
\usepackage{graphicx}
\usepackage{cite}
\usepackage{dsfont}
\usepackage{booktabs}
\usepackage{tabularx}
\usepackage{multirow}
\usepackage{amsmath}
\usepackage{enumitem}
\usepackage{latexsym}
\usepackage{eucal}
\usepackage{bm}
\usepackage{amsfonts}

\usepackage[switch]{lineno}

\usepackage{bbding}
\usepackage{pifont}

\usepackage{listings}
\usepackage{xcolor}
\usepackage{colortbl, xcolor} 
\lstdefinestyle{promptstyle}{
    language=bash,                
    basicstyle=\ttfamily\small,   
    keywordstyle=\color{blue},    
    stringstyle=\color{red},      
    commentstyle=\color{gray},    
    backgroundcolor=\color{gray!05}, 
    frame=tb,                     
    framerule=0.5pt,              
    rulecolor=\color{black!80},   
    breaklines=true,              
    postbreak=\mbox{\textcolor{red}{$\hookrightarrow$}\space}, 
    numbers=left,                 
    numberstyle=\tiny\color{gray},
    captionpos=b,                 
    escapeinside={(*@}{@*)}       
}

\usepackage{graphicx}       
\usepackage{ragged2e}       

\newcolumntype{L}[1]{>{\RaggedRight\arraybackslash}p{#1}}
\newcolumntype{Y}{>{\RaggedRight\arraybackslash}X}
\usepackage{arydshln}

\begin{document}

\title{From Hindsight to Foresight:\\Self-Encouraged Hindsight Distillation for \\Knowledge-based Visual Question Answering}
\author{Yu Zhao, 
Ying Zhang, 
Xuhui Sui, Baohang Zhou, Xinying Qian,
Li Shen, 
Dacheng Tao, \IEEEmembership{Fellow,~IEEE}
\thanks{Yu Zhao, Ying Zhang, Xuhui Sui, Xinying Qian are with the College of Computer Science, VCIP, DISSec, Nankai University, Tianjin, China (e-mail: zhaoyu@dbis.nankai.edu.cn, yingzhang@nankai.edu.cn, suixuhui@dbis.nankai.edu.cn, qianxinying@dbis.nankai.edu.cn).}
\thanks{Baohang Zhou is with the School of Software, Tiangong University, Tianjin, China (e-mail: zhoubaohang@tiangong.edu.cn).}
\thanks{Li Shen is with the Shenzhen Campus of Sun Yat-sen University, Shenzhen, China (e-mail: mathshenli@gmail.com).}
\thanks{Dacheng Tao is with the Generative AI Lab, College of Computing and Data Science, Nanyang Technological University, Singapore (e-mail: dacheng.tao@gmail.com).}
}




\maketitle


\begin{abstract}
Knowledge-based Visual Question Answering (KBVQA) necessitates external knowledge incorporation beyond cross-modal understanding. Existing KBVQA methods either utilize implicit knowledge in multimodal large language models via in-context learning or explicit knowledge via retrieval augmented generation. However, their reasoning processes remain implicit, without explicit multi-step trajectories. To address this gap, we propose a Self-Encouraged Hindsight Distillation Reasoning (HinD) framework, aiming at eliciting reasoning ability inside the MLLM by constructing a Hindsight Teacher with privileged information to teach the Foresight Student.
First, we construct the Hindsight Teacher by prompting the MLLM with the reasoning target as privileged information to complete the reasoning process, obtaining Hindsight-Zero training data. 
Then, the Foresight Student, without knowing the answer, learns the golden trajectories from Hindsight in two ways: (1) Hindsight Distillation Fine-Tuning to self-distill the Hindsight-Zero into a modularized Chain-of-Thought Generator and a Knowledge Generator for sequential steps and discrete facts generation, respectively; (2) Knowledge Encouragement Preference Optimization to encourage the under-confident but relevant knowledge inside the MLLM and suppress the over-confident but irrelevant one. 
Experiments on OK-VQA and A-OKVQA validate the effectiveness of HinD, showing that HinD with 7-8B MLLM achieves superior performance without commercial model APIs or retrieved knowledge.

\end{abstract}

\begin{IEEEkeywords}
Knowledge-based Visual Question Answering, Chain-of-Thought, Multimodal Large Language Models
\end{IEEEkeywords}

\section{Introduction}

\begin{figure}[!t]
\centering
\includegraphics[width=0.47\textwidth]{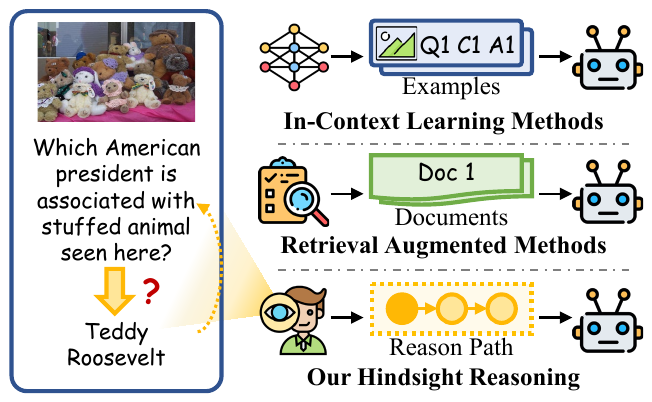}
\caption{KBVQA requires necessary knowledge incorporation for reasoning. Unlike existing in-context learning and retrieval-augmented methods, we elicit the reasoning ability inside MLLMs through Hindsight-distilled reasoning. }  \label{fig:intro}
\end{figure}

Knowledge-based Visual Question Answering (KBVQA) \cite{marino2019okvqa,schwenk2022aokvqa,DBLP:journals/tmm/KBVQA_Prompting,DBLP:journals/tmm/FactVQA} is a challenging task for cross-modal reasoning, aiming at answering the question based on the image and external knowledge. 
Unlike conventional VQA \cite{antol2015vqa,goyal2017vqav2}, which answers questions based solely on image content, KBVQA necessitates the commonsense incorporation beyond the image. 
Take Figure \ref{fig:intro} as an example, answering the question requires (1) identifying the stuffed animals in the image as \textit{teddy bears}, (2) associating the teddy bear with \textit{President Teddy Roosevelt}. 
Knowledge-based vision-language reasoning is essential for knowledge-intensive application scenarios \cite{shah2019kvqa,DBLP:journals/tmm/LuZS26KnowledgeDialog,DBLP:journals/tmm/WangLJWLLCL26KnowledgeVidelAction,DBLP:journals/tmm/OuyangTCZXXW26KnowledgeRecommendation}. 

Existing KBVQA studies can be categorized into two mainstreams: \textit{(1) {In-context learning methods}} \cite{hu2023promptcap,khademi2023mmreasoner,yu2025prophet,DBLP:journals/tmm/BiasKBVQA} (ICL) usually construct relative examples with captions \cite{yang2025QACap} to prompt large language models (LLMs) \cite{brown2020GPT3,achiam2023gpt4,hurst2024gpt4o,touvron2023llama2} to utilize their implicit knowledge. 
However, the ICL paradigm is basically an {imitation} of the context examples, lacking explicit reasoning paths to explain their reasonability. 
\textit{(2) {Retrieval augmented methods}} \cite{gui2022kat,lin2023FLMR,long2025reause,DBLP:journals/tmm/KBVQA_Prompting} usually retrieve relevant knowledge triples or documents to augment the answer prediction.
They usually fine-tune the Multimodal LLMs (MLLMs) to answer the question based on the retrieved knowledge, in which the reasoning source and process remain non-transparent. 

Inspired by Chain-of-Thought (CoT) \cite{wei2022CoT} studies, we propose to elicit the internal reasoning ability of MLLMs to provide step-by-step reasoning processes for KBVQA. 
CoT is widely explored in LLMs \cite{wei2022CoT,wang2022selfconsistency} and MLLMs \cite{wang2025multimodalCoTsurvey,DBLP:journals/tmm/FacialCoT,DBLP:journals/tmm/VisualCommonsenseCoT,DBLP:journals/tmm/MLLM26} to break down the reasoning process into human-like intermediate steps. 
It is promising for facilitating multi-hop inference in the KBVQA task that naturally requires multi-step reasoning. 
However, though the idea of CoT is intuitive, it is non-trivial for KBVQA task due to the following challenges:

\textbf{(1) \textit{Prohibitive Cost of Reasoning Supervision}.} 
Though KBVQA naturally requires multi-step reasoning for knowledge-informed answer prediction, most datasets do not include ground-truth systematic and structured reasoning trajectories. 
Existing multimodal CoT studies usually obtain CoT data with larger and more costly teacher MLLMs such as GPTs \cite{xu2024llava-cot,yao2024mulberry,jiang2025corvid} or manual construction \cite{schwenk2022aokvqa,lu2022ScienceQA}, which requires high computational or labor expenses.
This creates a significant bottleneck for constructing a KBVQA reasoning model, especially for smaller MLLMs. 
A natural remedy is to let the small MLLM generate the trajectories by itself, i.e., self-distillation \cite{yang2024selfdistillbridge,zelikman2022star}.
However, since KBVQA naturally requires knowledge beyond the vision-language input, the small MLLM can hardly anchor the critical knowledge without guidance, making the
self-generated trajectories largely imperfect as supervision.

\textbf{(2) \textit{Misalignment of Knowledge Correctness and Confidence}. }
KBVQA inherently demands associating knowledge outside the vision-language input. 
For example, \textit{Teddy Roosevelt} is crucial for reasoning in Figure \ref{fig:intro}, but did not appear in the image or the question of the sample.
However, smaller MLLMs are usually trained with visual-language aligned corpus.
It results in misalignment between generation confidence and correctness:
MLLMs are often over-confident when generating the vision-language concentrated contents that cannot help reasoning, such as \textit{Teddy bears are made from soft materials}; while under-confident when generating the helpful knowledge outside the vision-language input, such as \textit{Teddy bears are associated with Teddy Roosevelt}.
Thus, aligning correctness and confidence is crucial for knowledge reasoning of MLLMs. 

The two challenges boil down to one question: how to obtain supervision from a small MLLM that guides the MLLM on both what to generate and when to be confident. 
Inspired by human cognition, we note that humans are more capable of explaining and reconstructing reasoning paths towards a known conclusion in Hindsight than reasoning from scratch in Foresight. 
We hypothesize that the same asymmetry holds for MLLMs: conditioned on the target answer as an anchor, the MLLM produces more precise and higher-quality trajectories, which we validate in the Pilot Analysis.
Moreover, the Hindsight also tells when to be confident: it annotates where the confidence of the Foresight student is misaligned with the correctness. 
Therefore, by conditioning on the answer as privileged
information~\cite{vapnik2015learningusingprivilegedinformation,lopezpaz2016unifyingprivileged}, the small MLLM is turned into its own Hindsight teacher, from which a Foresight student learns from a forward perspective without answer access. 

In our paper, we propose the \textbf{Self-Encouraged Hindsight Distillation Reasoning (HinD) framework} for Knowledge-based Visual Question Answering, 
which uses a Hindsight teacher with privileged information to teach the Foresight student without it.
We construct the Hindsight-Zero training data as Hindsight teacher by directly prompting the MLLM to connect the intermediate dots between the input and the reasoning target, which enables the generation of goal-oriented and conclusion-aligned reasoning data from small MLLMs. 
Since sequential reasoning steps and discrete relevant facts follow different semantic distributions, we distill the Hindsight-Zero to fine-tune the Foresight Student, composed of CoT Generator and Knowledge Generator. The modularized design enables clean distribution alignment for easier optimization. 
To tackle the knowledge confidence-correctness misalignment, the Hindsight teacher annotates and guides the Foresight student by verifying when the student should be confident. 
Specifically, we propose Knowledge Encouragement Preference Optimization to optimize the Knowledge Generator to encourage the generation of under-confident but relevant knowledge over the over-confident but unhelpful ones.
Finally, Answer Generator summarizes an answer based on the generated CoT and sampled knowledge. 
Experiments on OK-VQA and A-OKVQA demonstrate the effectiveness of HinD. 

Our contributions can be summarized as follows:
\begin{itemize}
    \item We introduce HinD, a Self-Encouraged Hindsight Distillation Reasoning framework that enables hindsight-to-foresight self-teaching by constructing answer-conditioned Hindsight-Zero trajectories as self-generated supervision.
    \item The Hindsight Teacher with privileged information serves as the trajectory source of Hindsight Fine-tuning, and confidence-correctness alignment annotator of Knowledge Encouragement Preference Optimization.
    \item Experiments on OK-VQA and A-OKVQA show that HinD achieves superior performance without GPT APIs or retrieved knowledge. The PRR@K of 96.7 shows HinD with MLLM parameter knowledge provides higher relevant knowledge than the retrieval-augmented methods. 
\end{itemize}

\section{Related Work}

\subsection{Knowledge-based VQA}

Visual Question Answering (VQA) \cite{antol2015vqa,goyal2017vqav2} is a crucial task for cross-modal reasoning aiming at answering visual-related questions, such as colors, counting, etc.
To address the VQA challenges when visual information is not sufficient for reasoning and the question needs the necessary outside knowledge to answer, the Knowledge-based VQA task is proposed \cite{marino2019okvqa,schwenk2022aokvqa,chen2023infoseek}, aiming at reasoning with cross-modal understanding and necessary knowledge incorporation. 
Existing approaches can be categorized into two mainstreams:

\textbf{(1) In-Context Learning Methods. }
These methods \cite{yang2022PICa,hu2023promptcap,yu2025prophet,khademi2023mmreasoner,xenos2023SimKBVQA} usually exploit the implicit knowledge from large language models, such as GPT-3 \cite{achiam2023gpt4}, GPT-4 \cite{achiam2023gpt4, hurst2024gpt4o}, Llama \cite{touvron2023llama2}, etc. 
They usually construct in-context learning (ICL) examples and follow the few-shot learning paradigm to output the answer. 
Since the prompted LLMs are usually language-only models, they usually construct image captions \cite{yang2022PICa,hu2023promptcap}, object labels \cite{khademi2023mmreasoner}, scene-graphs \cite{chen2024visualCoT-kbvqa}, OCRs \cite{khademi2023mmreasoner}, etc, to turn visual modality into textual descriptions to enable cross-modal understanding \cite{hu2023promptcap}. 
The in-context examples are selected through cross-modal similarity \cite{yang2022PICa} or a smaller VQA model \cite{yu2025prophet,xenos2023SimKBVQA}. 
However, the ICL methods are imitations of demonstrated examples, while the explicit reasoning sources and processes are unavailable. 

\textbf{(2) Retrieval Augmented Methods. }
These methods usually incorporate explicit outside knowledge from Knowledge Graphs such as Concept-Net \cite{speer2017conceptnet, garderes2020conceptbert,dong2024MAIL}
 and Wikidata \cite{vrandevcic2014wikidata,gui2022kat,lin2022revive};
document corpora such as Wikipedia \cite{marino2019okvqa,marino2021krisp,wu2022mavex,gao2022trig,si2023TwO} and Google Search \cite{luo2021VisDPR,lin2022RAVQA,lin2023FLMR,long2025reause,Fang_2025_CVPR_NoteMR}; 
multimodal knowledge such as VQA dataset \cite{si2023TwO}, VisualGenome \cite{marino2021krisp}, and Google Images \cite{wu2022mavex}. 
They prompt the retrieved knowledge to GPT APIs \cite{gui2022kat,lin2022revive,si2023TwO} or fine-tune an MLLM \cite{li2023blip2,raffel2020T5,dai2023instructblip} as the answer generator. 
However, the retrieved knowledge noises \cite{yu2025prophet} limit the KBVQA performance, which also lacks further exploitation of MLLMs' internal reasoning ability and explicit reasoning processes.

\subsection{Multimodal Chain-of-Thought}
Recently, LLMs \cite{brown2020GPT3,achiam2023gpt4,hurst2024gpt4o,guo2025deepseekR1} have demonstrated strong reasoning ability, exploiting the Chain-of-Thought (CoT) \cite{wei2022CoT} to elicit the human-alike intermediate step-by-step reasoning procedures. 
The Multimodal CoT (MCoT) \cite{wang2025multimodalCoTsurvey,xu2024llava-cot,zhang2024TMLR-MM-CoT,chen2024visualCoT-kbvqa,DBLP:journals/tmm/FacialCoT,DBLP:journals/tmm/VisualCommonsenseCoT} studies further extend CoT into cooperations of multiple modalities, including images, videos, audios, etc. 
The CoT Prompting strategy can be applied to Multimodal Large Language Models (MLLMs) such as GPT-4o \cite{hurst2024gpt4o} following zero-shot or ICL paradigms to obtain the reasoning steps for improving reasoning accuracy. 
Existing MCoT KBVQA studies \cite{chen2024visualCoT-kbvqa,zhang2024TMLR-MM-CoT} follow the ICL paradigm with captions, scene-graphs, rationales, etc, as contexts, without exploring step-by-step trajectories and internal knowledge generation in our work. 
Our work differs from theirs in proposing a CoT-tuning framework to fully elicit and exploit MLLM's internal reasoning ability. 
Moreover, since there are usually no reasoning data for common datasets, CoT supervision data is often from human annotations \cite{schwenk2022aokvqa,lu2022ScienceQA} or larger LLM guidance \cite{yao2024mulberry,xu2024llava-cot,jiang2025corvid}, which are usually computationally expensive. 
On the contrary, we propose to exploit the Hindsight wisdom in MLLMs to construct reasoning data for self-taught distillation, decreasing tuning difficulty and data construction expenses.

\begin{figure*}[t]
\centering
\includegraphics[width=\textwidth]{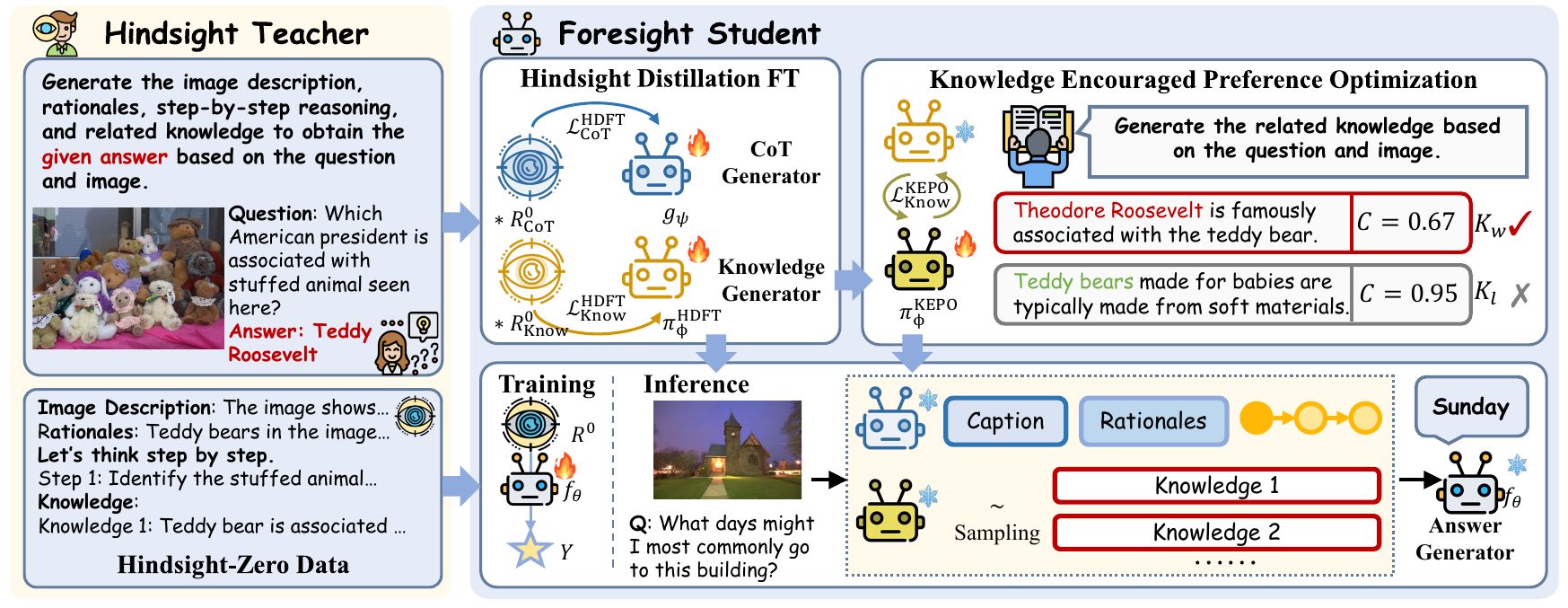}
\caption{Our Self-Encouraged Hindsight Distillation Reasoning (HinD) framework for KBVQA.
\textit{(1) Hindsight Teacher} is constructed by completing intermediate reasoning processes from question to ground truth answer, forming Hindsight-Zero data.
\textit{(2) Foresight Student} is trained through: \textit{Hindsight Distillation Fine-Tuning} for CoT Generator and Knowledge Generator; 
\textit{Knowledge Encouragement Preference Optimization} for aligning confidence and correctness. 
Finally, the Answer Generator summarizes the answer.} \label{fig:method}
\end{figure*}

\section{Formulation and Pilot Analysis}

\subsection{Task Formulation. }
Knowledge-based Visual Question Answering (KBVQA) task aims to answer the question based on the image input and outside knowledge. 
Formally, we define the input sample as $X = (Q, I)$ and its target answer as $Y$. 
The reasoning model $f_\theta(\cdot)$, often an LLM or MLLM, aims to answer the question based on the given context $C$, as $\hat{Y} = f_\theta (X, C)$.

\subsection{Paradigm Comparisons. }
For the In-Context Learning methods, their contexts are selected examples $C_\text{context}=\mathcal{E}=\{(X_i, Y_i)\}$, where $\mathcal{E}$ denotes example set. 
For the Retrieval Augmented methods, their contexts are retrieved documents or knowledge triples $C_\text{retrieval}=\mathcal{D} = \{D_i\}$, where $\mathcal{D}$ denotes document/knowledge set. 
For HinD, our contexts are the intermediate reasoning processes including the Chain-of-Thoughts (CoT) and Knowledge $C_\text{HinD} = (\text{IC, RAT}, \mathcal{S}, \mathcal{K})$, which denotes image caption (IC), rationales (RAT), sequential reasoning steps $\mathcal{S}=[S_i]$, and discrete knowledge pieces $\mathcal{K}=\{K_i\}$.
For convenience, we jointly denote the $(\text{IC, RAT},\mathcal{S})$ as CoT.

\subsection{Hindsight and Foresight Formulation. }
The traditional reasoning process paradigm with LLMs is as $f: X \rightarrow R, Y$, where the reasoning path $R$ and answer $Y$ are from only the input sample $X$.
In terms of \textit{Hindsight Teacher}, we obtain the golden trajectory $R$ by prompting both the input and its target answer as $f: X \times Y \rightarrow R$, which are contextually aligned with the input and also goal-targeted to the ground truth answer. 
In implementation, we fix the MLLM and prompt the training set data, in which the answers serve as available guidance, to construct the structured reasoning process  $R^0 = f(R|
X, Y) = (\text{IC}^0, \text{RAT}^0, \mathcal{S}^0, \mathcal{K}^0) = (R^0_\text{CoT}, R^0_\text{Know})$ for subsequent distillation, which we denote as \textbf{Hindsight-Zero}.
In terms of \textit{Foresight Student}, it has no access to answer and follows the traditional reasoning paradigm as $f: X \rightarrow R, Y$. In implementation, we modularized the CoT and Knowledge generation, thus the reasoning is separated into two stages: $f: X \rightarrow R$ and $f: X, R \rightarrow Y$. In the whole process, the answer is unavailable for Foresight Student to avoid leakage and potential shortcuts. 

\subsection{Pilot Analysis. }
To investigate whether the answer-conditioned Hindsight yields more precise reasoning data, and whether its reasoning quality is sufficient to support subsequent distillation, we conduct the following pilot analysis on the Hindsight-Zero.

\subsubsection{Quantitative analysis}
\begin{table}[t]
\caption{Pilot Analysis between Hindsight-Zero (HZ) and Self-Distillation (SD). }\label{tab:pilot-SD-HD}
\centering\small
\begin{tabular}{ccccc}
\toprule
Train&SD-$R^0_\text{CoT}$ & SD-$R^0_\text{Know}$ & HZ-$R^0_\text{CoT}$ & HZ-$R^0_\text{Know}$ \\
\midrule
PRR@K&73.9 & 81.0 & 85.4 & 98.0 \\
\bottomrule
\end{tabular}
\end{table}

We compare Hindsight-Zero (HZ) with a matched Self-Distillation (SD) baseline
sampled from $f(R|X)$ with the identical MLLM, prompt, and filtering, differing
only in whether the answer $Y$ is provided. As in Table~\ref{tab:pilot-SD-HD}, {HZ lifts PRR@K by 11.5 on CoT and 17.0 on Knowledge}.
It is reasonable for the KBVQA nature, since the critical
entity in KBVQA usually relates to key entities outside the image and question, which is challenging for a small MLLM to anchor without guidance. It confirms that {the asymmetric information of Hindsight is necessary for constructing usable supervision for KBVQA from small MLLMs}.

\subsubsection{Qualitative analysis}
\begin{table}[t]
\caption{Pilot Analysis of Hindsight-Zero Quality Evaluation. }\label{tab:pilot-quality}
\centering\small
\begin{tabular}{cccc}
\toprule
Grounded & Usefulness & CoT Logic & Knowledge Quality \\
\midrule
4.52 & 4.24 & 4.70 & 4.16\\
\bottomrule
\end{tabular}
\end{table}
We employ DeepSeek-V4-Flash \cite{deepseek2026v4}
as an LLM-Judge to rate 100 random Hindsight-Zero samples (1 to 5) on groundedness, usefulness, CoT logic, and knowledge quality. As in
Table~\ref{tab:pilot-quality}, {all four dimensions score above 4.1}, indicating that {Hindsight-Zero provides grounded and logically coherent trajectories with a small MLLM, qualified as golden supervision. 

\section{Method}
Figure \ref{fig:method} shows the Self-Encouraged Hindsight Distillation Reasoning framework. 
Validated by the pilot analysis, we take Hindsight-Zero $R^0$ as golden trajectories and distill it into the Foresight Student, which has no access
to the answer, via:
(1) Hindsight Fine-tuning into CoT Generator and Knowledge Generator. 
(2) Knowledge Encouragement Preference optimization to align knowledge generation correctness and confidence.
Finally, the Answer Generator summarizes the answer from the generated CoT and knowledge.

\subsection{Hindsight Distillation Fine-tuning}
After obtaining Hindsight-Zero $R^0 = f(R|
X, Y)$, we distill Hindsight-Zero to the MLLM through Hindsight Distillation Fine-Tuning (HDFT). 
This way, the tuned MLLMs can generate the reasoning process from input samples as $f: X \rightarrow R$, guided to generate the idealized reasoning process without seeing the answer. 
By distilling Hindsight-Zero, the reasoning model could generalize to test samples through a wiser perspective since they have seen the right move towards the right answer through HDFT. 

\textbf{Hindsight Self-Distillation. }
We exploit the same MLLM backbone as the one producing $R^0$. 
Learning from the data generated by itself could align with the model distributions to avoid huge tuning offset \cite{yang2024selfdistillbridge}, which lowers the fine-tuning difficulty. 
Moreover, we divide the reasoning process generation into CoT Generator and Knowledge Generator, since CoT is a step-by-step sequential process, while the knowledge generation requires the model to recall the discrete related facts. 

\textbf{Quality Control. }
Since reasoning trajectories are free-form natural languages, it is hard to automatically verify whether a certain trajectory supports the reasoning.
Inspired by the pseudo-relevance notion in KBVQA \cite{lin2023FLMR}, we use answer coverage as a simple deterministic acceptance criterion and retain only trajectories ${^*R^{0}}$ containing $Y$. 
This criterion does not imply that answer-missing trajectories are irrelevant, but only selects supervision whose support for the target answer can be automatically verified. 
The answer $Y$ is used solely to generate and filter teacher trajectories from the training split and is never provided to the Foresight Student or accessed from the validation and test splits, preventing leakage. This procedure serves as rejection-style data filtering \cite{guo2025deepseekR1}.

\textbf{CoT HDFT. }
The CoT data is defined as $R_\text{CoT} = (\text{IC,RAT},\mathcal{S})$. 
By prompting the MLLMs to generate the image caption, rationales, and step-by-step reasoning according to the input question and image, we fine-tune the CoT Generator to output the Hindsight-Zero data $^*R^0_\text{CoT}$. 
The training objective is given in Equation \eqref{eq: hdft cot}, where $g_{\psi}$ denotes the CoT Generator.
\begin{equation} \label{eq: hdft cot}
\small
    \mathcal{L}_\text{HDFT}^\text{CoT} = - \log g_{\psi}({^*R^0_\text{CoT}}| X)
\end{equation}

\textbf{Knowledge HDFT. }
In $R^0_\text{Know}$ generation, we prompt the fixed MLLM to output a fixed number of related knowledge or facts. Since known facts are usually discrete, we split the Knowledge Generator from the whole reasoning process, and fine-tune it to generate a single piece of knowledge. 
After obtaining the knowledge set $\mathcal{K}^0$ in Hindsight-Zero, we split the knowledge pieces into hit or missed ground truth answer, as $\mathcal{K}^0 = \mathcal{K}^0_\text{hit} \bigcup \mathcal{K}^0_\text{miss}$. 
For fine-tuning, we randomly select one of the hit knowledge $^*R_\text{Know}^0 \sim \mathcal{U}(\mathcal{K}_\text{hit})$ as the training objective.
The objective is in Equation \eqref{eq:hdft know}, where $\pi_{\phi}^\text{HDFT}$ is the fine-tuned Knowledge Generator. 
\begin{equation} \label{eq:hdft know}
\small
    \mathcal{L}_\text{HDFT}^\text{Know} = - \log \pi_{\phi}^\text{HDFT}(^*R_\text{Know}^0 | X)
\end{equation}

\subsection{Knowledge Encouragement Preference Optimization}

To further align the knowledge correctness and confidence, we propose a Knowledge Encouragement Preference Optimization (KEPO) for the Knowledge Generator.  
The misalignment reason could be that KBVQA task requires circuitous knowledge outside the vision-language input to answer the question, rather than those directly appearing in the question or image.
In our paper, we propose to encourage the less-confident but helpful knowledge, while rejecting the over-confident but unhelpful one. 
In this way, the MLLM is calibrated to learn when to be more assured. 

\textbf{Self-Encouragement Pair Construction. }
We first sample a set of knowledge $\mathcal{K}$ from the Knowledge Generator after HDFT $\pi^\text{HDFT}_\phi(\cdot)$ as the reference model. The self-sampled pairs lower the KEPO difficulty. 
We calculate the confidence of knowledge generation sample $K$ as Equation \eqref{eq: kgen conf}, which is the normalized generation likelihood of the reference model. 
The $T$ denotes number of tokens of $K$, and $p_{\pi_{ref}}(K_t|K_{<t})$ denotes conditional probability for $t$-th token. 
\begin{equation} \label{eq: kgen conf}
\small
C(K) = \exp \left( \frac{\sum_1^T \log p_{\pi_{ref}}(K_t|K_{<t})}{T}\right)
\end{equation}

We construct the (preferred, dispreferred) pairs for KEPO as $(K_w, K_l)\sim \pi^\text{HDFT}_\phi(K|X)$, which are defined as Equation \eqref{eq: kepo pair}, respectively. 
The $K_w$ is the winning knowledge with the \textit{minimum} confidence in $\mathcal{K}_\text{hit}$, while the $K_l$ is the losing knowledge with the \textit{maximum} confidence in $\mathcal{K}_\text{miss}$. 
\begin{equation} \label{eq: kepo pair}
\small
\begin{cases}
K_w = \underset{K\in\mathcal{K}_\text{hit}}{\arg \min}~C(K) \\
K_l = \underset{K\in\mathcal{K}_\text{miss}}{\arg \max}~C(K) 
\end{cases}
\end{equation}

\textbf{Optimization Objective. } KEPO exploits DPO \cite{rafailov2023DPO} loss to guide MLLM to distinguish the constructed pair as Equation \eqref{eq: kepo}. 
The $\pi_\phi^\text{KEPO}$ and $\pi_\text{ref}$ are both initialized to HDFT optimized Knowledge Generator $\pi_{\phi}^\text{HDFT}$. 
The $\pi_\phi^\text{KEPO}$ parameters are optimized, while the $\pi_\text{ref}$ is the fixed reference model from which the pairs are sampled and annotated from hindsight.
\begin{equation} \label{eq: kepo}
\small
    \mathcal{L}_\text{KEPO} =
    -  \log \sigma \left( \beta \log \frac{\pi_\phi^\text{KEPO} (K_w|X)}{\pi_\text{ref}(K_w|X)} - \log \frac{\pi_\phi^\text{KEPO}(K_l|X)}{\pi_\text{ref}(K_l|X)}\right)
\end{equation}

\subsection{Answer Generation}
\textbf{Training. }
We fine-tune the Answer Generator to concisely summarize the answer with the given image, question, and reasoning process as Equation \eqref{eq: ans gen}, in which the $f_\theta(\cdot)$ is the optimized Answer Generator. 
In implementation, CoT and Knowledge that do not contain the ground-truth answer are preserved, and knowledge pieces are shuffled. 
This way, the generalization of the Answer Generator is improved in case the generated CoT and knowledge do not include the answer. 
\begin{equation}\label{eq: ans gen}
\small
\mathcal{L}_\text{ans} = - \log f_{\theta}(Y | X, R_\text{CoT}^0, \mathcal{K}^0)
\end{equation}

\textbf{Inference. }
In the test phase, given the test sample $X$, we obtain the CoT $\hat{R}_\text{CoT} = g_\psi (R_\text{CoT}|X)$ and sampled knowledge $\hat{\mathcal{K}} = \{\hat{R}_\text{Know}^i\} \sim \pi_\phi^\text{KEPO} (R_\text{Know}|X)$. 
Then, the answer is predicted as $\hat{Y} = f_\theta (Y| X, \hat{R}_\text{CoT}, \hat{\mathcal{K}})$.
We perform self-consistency (SC) \cite{wang2022selfconsistency} with major-vote on 5 runs with different seeds for reasoning robustness following \cite{yu2025prophet,hu2023promptcap}. 

\section{Experiments}

\subsection{Experimental Settings}

\textbf{Dataset. }
We conduct experiments on two widely-adopted KBVQA benchmarks, OK-VQA \cite{marino2019okvqa} and A-OKVQA \cite{schwenk2022aokvqa}.
OK-VQA is a commonly used KBVQA dataset, consisting of 9K/5K image-question-answer samples for train/test. 
A-OKVQA is a large KBVQA dataset which requires more reasoning and commonsense knowledge, with 17K/1K samples for train/valid.
We also conduct generalization validation on ScienceQA \cite{lu2022ScienceQA} and MME \cite{fu2026mme} datasets.

\textbf{Evaluation Metrics. }
OK-VQA is evaluated under the direct answering (DA) setting, while A-OKVQA provides both DA and multiple-choice (MC) settings. 
For OK-VQA and A-OKVQA (DA), we follow previous studies and exploit the VQA score \cite{marino2019okvqa} to evaluate the prediction accuracy. 
OK-VQA and A-OKVQA (DA) provide 10 ground truth answers $\mathcal{A}$ for each sample. 
The VQA score is a soft accuracy based on the occurrence count in the annotated answers: VQA Score$(\hat{Y}, \mathcal{A})=\min (\#\mathcal{A}(\hat{Y})/3, 1)$. 
For A-OKVQA (MC), multi-choice accuracy is reported. 
Following retrieval augmented KBVQA studies \cite{luo2021VisDPR,lin2023FLMR,long2025reause}, we report Pseudo-relevance Recall@K (PRR@K) \cite{luo2021VisDPR} to evaluate whether K pieces of knowledge (and 1 CoT sample for HinD-CoT-Know) contain any ground-truth answers. 
We also extract recalls of selected examples \cite{yu2025prophet,chen2024visualCoT-kbvqa} from in-context learning methods. 

\textbf{Baselines. }
We compare HinD with three groups of baselines: \textit{(1) MLLM baselines}: LLaVA \cite{liu2023llava}, MiniGPT4 \cite{zhu2023minigpt4}, Qwen2.5 \cite{team2024qwen2report}, PaLM-E \cite{driess2023palme}, GPT-4V \cite{achiam2023gpt4}. 
\textit{(2) In-context Learning Methods}: PromptCap \cite{hu2023promptcap}, VisualCoT \cite{chen2024visualCoT-kbvqa}, MinBias \cite{DBLP:journals/tmm/BiasKBVQA}, Prophet and Prophet++ \cite{yu2025prophet}, MM-Reasoner \cite{khademi2023mmreasoner}, SimKBVQA \cite{xenos2023SimKBVQA}, MM-CoT \cite{zhang2024TMLR-MM-CoT}, LSMS \cite{sun2025LSMS25MM}, QACap \cite{yang2025QACap}.
\textit{(3) Retrieved Augmented Methods: }
KAT \cite{gui2022kat}, ReVIVE \cite{lin2022revive}, TwO \cite{si2023TwO}, TRiG \cite{gao2022trig}, RA-VQA \cite{lin2022RAVQA}, ReVeal \cite{hu2023reveal}, MAIL \cite{dong2024MAIL}, FLMR \cite{lin2023FLMR}, MMRP \cite{DBLP:journals/tmm/KBVQA_Prompting}, Boter \cite{hao2024boter}, SKP \cite{wang2024SKP24ACL}, LLM-RA \cite{jian2024LLMRA}, NoteMR \cite{Fang_2025_CVPR_NoteMR}, ReAuSE \cite{long2025reause}. 
We compare HinD-CoT-Know and HinD-Know to the baselines, in which HinD-Know refers to HinD with only generated knowledge for comparison with retrieval-augmented baselines.

\textbf{Implementation details. }
In this paper, we exploit Qwen2.5-VL (7B) \cite{team2024qwen2report} as backbone MLLM. We also report results of other MLLMs distilled from Qwen2.5-VL-7B Hindsight-Zero data in Table \ref{tab: OKVQA} and Table \ref{tab:backbone}.
We freeze the parameters of MLLM and fine-tune LoRA \cite{hu2022lora} parameters of CoT Generator, Knowledge Generator, and Answer Generator. 
Knowledge Generator requires two sets of LoRA for HDFT and KEPO. 
We exploit LLaMA-Factory \cite{zheng2024llamafactory} for efficient tuning. 
Knowledge is sampled with default parameters: temperature=1.0, top-p=0.95, top-k=50. 
We set $|\mathcal{\hat{K}}|$=5 to mimic the 5 documents in the retrieval-augmented baselines. 
Answer generation follows low-temperature (1e-5) configs for improving reasoning certainty. 
Experiments are conducted with 1-2 A6000 GPUs with 48 GB memory. 
\textit{More details and discussions are provided in Supplementary Material.}

\begin{table}[!t]
\caption{KBVQA performance on OK-VQA dataset. WD, W, GS, and C refer to Wikidata, Wikipedia passages, Google Search documents, and Concept-Net, respectively. $^\dagger$ denotes K knowledge pieces plus 1 CoT process. }
\label{tab: OKVQA}
\vspace{-10pt}
\begin{center}
\resizebox{\linewidth}{!}{
\centering
\small
\begin{tabular}{lccc}
\toprule
Model & Knowledge Source & PRR@K & Score  \\
       \midrule
\multicolumn{4}{c}{\textit{\textbf{Multimodal Large Language Models}}} \\
LLaVA-13B           & - & - & 61.9 \\
Minigpt4-v2-7B      & - & - & 57.8 \\
Minigpt4-v2-7B (FT) & - & - & 61.9 \\
Qwen2.5-VL-7B       & - & - & 61.6 \\
Qwen2.5-VL-7B (FT)  & - & - & 62.0 \\
PaLM-E-562B         & - & - & 66.1 \\
GPT-4V              & - & - & 64.3 \\
\midrule
\multicolumn{4}{c}{\textit{\textbf{In-Context Learning Methods}}} \\
PromptCap   & GPT-3        & - & 60.4 \\
VisualCoT & BLIP2+Llama-2-70B & 66.0 & 54.9 \\
MinBias & Llama-2-13B & - & 60.4 \\
Prophet     & GPT-3     & 79.8 & 61.1 \\
Prophet++     & GPT-4o       & 79.8 & 65.7 \\
MM-Reasoner & GPT-4        & - & 60.8 \\
SimKBVQA    & Llama2-13B & - & 61.2 \\
LSMS   &  LLaVA-1.5-13B                   &    -  & 65.0 \\
QACap & Claude 3.5 & - & 68.2 \\
\midrule
\multicolumn{4}{c}{\textit{\textbf{Retrieval Augmented Methods}}} \\
KAT    & WD+GPT-3             & -    & 54.4 \\
ReVIVE & WD+GPT-3             & -    & 58.0 \\
TwO    & W+VQAv2+GPT-3       & -    & 56.7 \\
TRiG   & W+T5                & 45.5 & 50.5 \\
RA-VQA & GS+T5               & 82.8 & 54.5 \\
ReVeaL & W+T5                & -    & 59.1 \\
MAIL   & C+InstructBlip      & -    & 61.8 \\
FLMR   & GS+BLIP2+T5-XL (3B) & 89.3 & 62.1 \\
MMRP & GS+WD+GPT-4o-mini & - & 62.2 \\
Boter  & GS+BLIP2+T5-XL (3B) & -    & 62.8 \\
SKP    & GS+Vicuna (7B) & - & 63.3 \\
LLM-RA & GS+BLIP2+T5-XL (3B) & 90.4 & 63.3 \\
NoteMR & GS+Qwen2-VL-7B & - & 64.8 \\
ReAuSE & GS+Minigpt4-7B    & 92.6 & 65.7 \\
\midrule
\rowcolor{blue!10} \textbf{HinD-Know}  & Qwen2.5-VL-7B    &{94.7}      & {68.3}      \\
\rowcolor{blue!10}\textbf{HinD-CoT-Know}  & Qwen2.5-VL-7B  & {95.3}$^\dagger$ &     {67.5} \\
\midrule
\rowcolor{blue!5} \multicolumn{4}{l}{\textit{\textbf{Distillation from Qwen2.5-VL-7B Hindsight-Zero}}}\\

\rowcolor{blue!5}\textbf{HinD-Know}  & Qwen2.5-VL-3B    &    94.8 & 64.7   \\
\rowcolor{blue!5}\textbf{HinD-CoT-Know}  & Qwen2.5-VL-3B  &  95.6$^\dagger$ & 65.9\\

\rowcolor{blue!5}\textbf{HinD-Know}  & InternVL3-8B    & \underline{95.8}      & \underline{68.4}      \\
\rowcolor{blue!5}\textbf{HinD-CoT-Know}  & InternVL3-8B  & \textbf{96.7}$^\dagger$ &     \textbf{68.6} \\
\bottomrule
\end{tabular}}

\vspace{-15pt}
\end{center}
\end{table}

\subsection{Main Results}

Table \ref{tab: OKVQA} and Table \ref{tab: AOKVQA} show the performance of HinD and baselines on OK-VQA and A-OKVQA datasets. 

 \textbf{$\bullet$ Comparison with MLLM baselines: }
On OK-VQA, our HinD-Know achieves 68.3 VQA score, surpassing even 562B-parameter PaLM-E by 2.2\% and GPT-4V by 4.0\% while using only a 7B backbone. 
On A-OKVQA, HinD-CoT-Know attains {87.2} MC accuracy and {69.0} DA accuracy, outperforming fine-tuned MLLMs like MiniGPT4-v2 and Qwen2.5-VL. 
HinD-CoT-Know and HinD-Know outperform Qwen2.5-VL (FT), demonstrating the effectiveness of HinD in exploiting the explicit reasoning process. 

 \textbf{$\bullet$ Comparison with In-Context Learning Methods: }
On OK-VQA, HinD-Know exceeds Prophet++ based on GPT-4o by 2.6\% and reaches comparable results with QACap based on Claude 3.5, demonstrating effectiveness. 
Comparing PRR@K, 95.3\% samples in HinD-CoT-Know contain ground-truth answers, which outperforms selected examples in VisualCoT and Prophet by 29.3\% and 15.5\%.
It shows that HinD could generate more valuable contexts through hindsight distillation. 
On A-OKVQA-MC, HinD-CoT-Know outperforms QACap by 10.5\%.
Moreover, MM-CoT, VisualCoT, and SimKBVQA cannot apply to the MC setting limited by in-context learning paradigm, while HinD is not constrained. 
\begin{figure*}[!t]
\centering
\includegraphics[width=\textwidth]{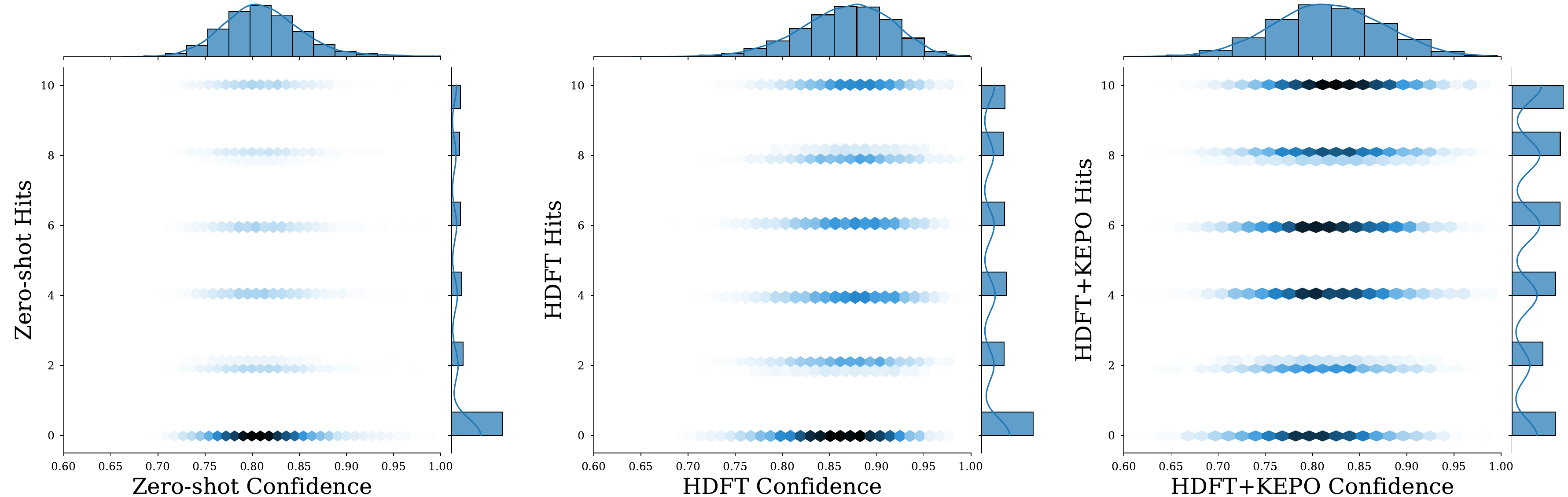}
\caption{The joint distribution of Knowledge Generator's confidence $C$ and Hits (mentioned answer count out of 10 annotated ground truth answers) on OK-VQA test set: (a) Zero-shot model (Qwen2.5-VL-7B), (b) HinD-Know with HDFT, and (c) HinD-Know with HDFT and KEPO.} \label{fig:kgen_hits_conf}
\end{figure*}

\begin{table}[!t]
\caption{KBVQA performance on A-OKVQA dataset. }
\label{tab: AOKVQA}
\vspace{-10pt}
\begin{center}
\resizebox{0.85\linewidth}{!}{
\centering
\small

\begin{tabular}{l c c}
\toprule
Model & Multi-Choice & Direct-Answer \\

       \midrule

LLaVA-1.5-7B         & 77.1 & 63.7 \\
InstructBLIP-7B (FT) & 73.0 & 62.4 \\
Minigpt4-v2-7B (FT)  & -    & 61.3 \\
Qwen2-VL 7B (FT)     & -    & 61.5 \\
GPV-2                & 60.3 & 48.6 \\
\midrule
MM-CoT & - & 50.6 \\
VisualCoT & - &  50.5  \\ 
PromptCap            & 73.2 & 56.3 \\
Prophet              & 76.4 & 58.2 \\
SimKBVQA               & -    & 58.6 \\
QACap & 76.7 &66.3  \\

\midrule
REVEAL               & -    & 52.2 \\
ReAuSE               & 85.0 & 67.7 \\
\midrule
\rowcolor{blue!10}\textbf{HinD-Know}     & \underline{85.9} & \underline{67.8} \\
\rowcolor{blue!10}\textbf{HinD-CoT-Know} & \textbf{87.2} & \textbf{69.0}  \\
\bottomrule
\end{tabular}}
\end{center}
\end{table}

 \textbf{$\bullet$ Comparison with Retrieval Augmented Methods: }
On OK-VQA, HinD-Know outperforms ReAuSE by 2.1\% in PRR@K and 2.6\% in VQA score, based on the same 7B-sized MLLM and without outside knowledge.
On A-OKVQA, HinD-CoT-Know outperforms ReAuSE 2.2\% MC accuracy and 1.3\% DA accuracy. 
The performance improvements of HinD-Know demonstrate that sufficient knowledge is elicited from MLLM parameters through HinD reasoning, providing more relevant knowledge than retrieval-based methods.

 \textbf{$\bullet$ Comparison between HinD-CoT-Know and HinD-Know:} HinD-Know performs better on OK-VQA even though HinD-CoT-Know has a higher PRR@K, while HinD-CoT-Know performs better on A-OKVQA. 
The reason could be due to MLLM characteristics, since the HinD-CoT-Know on InternVL3-8B shows better performance than HinD-Know.

\subsection{Ablation Study}
Ablation study on OK-VQA is in Table \ref{tab: ablation}. 

 \textbf{$\bullet$ Architectural Design:}
\textit{Comparing HinD-CoT-Know, HinD-Know, and HinD-CoT}, it is obvious that the PRR@K improvements are mainly from generated knowledge. 
\textit{By ablating self-consistency (SC)}, the average accuracy of 5 runs decreases by 1.5 with standard deviation 0.2, demonstrating that SC could bring robustness and improve reasoning performance. 
\textit{Ablating the modularized design}, the joint Knowledge-CoT decreases the PRR@K by 14.6\%, demonstrating the effectiveness of modularized Knowledge and CoT generator; the joint CoT-Answer generator decreases the VQA Score by 4.2\%, demonstrating the necessity of decoupled answer generator.

 \textbf{$\bullet$ Effect of HDFT and KEPO: }
\textit{Comparing zero-shot HinD-CoT (w/o HDFT) and HinD-Know (w/o HDFT w/o KEPO)}, the zero-shot generated knowledge (80.6) includes more ground-truth answers than CoT (71.4), which could serve as a heuristic for KBVQA that facts provide more relevant knowledge than reasoning steps.
\textit{By ablating confidence in KEPO pair construction}, the recall of HinD-Know decreases by 4.8\%.
It demonstrates that by calibrating the Knowledge Generator with self-confidence from its hindsight distillation fine-tuned version, the model is precisely encouraged to generate accurate knowledge. 
\textit{By ablating KEPO and HDFT}, PRR@K and VQA Score are both decreased, demonstrating the effectiveness of our proposed KEPO and HDFT.

\begin{table}[!t]
\caption{Ablation Study on OK-VQA. Decouplement ablation denotes the joint generation of Knowledge-CoT and CoT-Answer. w/o $C$ in KEPO denotes randomly constructing the hit and miss knowledge pair without considering confidence. }
\label{tab: ablation}
\begin{center}
\resizebox{0.9\linewidth}{!}{
\centering
\small
\begin{tabular}{lcc}
\toprule
Model & PRR@K & Score  \\
       \midrule
\rowcolor{blue!10}HinD-CoT-Know  & {95.3} &     67.5 \\
\quad w/o Decoupled Knowledge-CoT & 80.7 & 63.1 \\
\quad w/o Decoupled CoT-Answerer & 94.6  & 63.3 \\
\quad w/o Self-Consistency & - & 66.0 $^{\pm 0.2}$ \\
\midrule
\rowcolor{blue!10}HinD-CoT & 75.7 & 64.0 \\
\quad w/o HDFT & 71.4 & 60.6 \\
\midrule
\rowcolor{blue!10}HinD-Know & {94.7}      & 68.3      \\
\quad w/o $C$ in KEPO & 89.9 & 65.3 \\
\quad w/o KEPO         & 85.0 & 63.5 \\
\quad w/o KEPO w/o HDFT     & 80.6 & 62.9 \\

\bottomrule
\end{tabular}}

\end{center}
\end{table}

\begin{figure*}[!t]
\centering
\includegraphics[width=\textwidth]{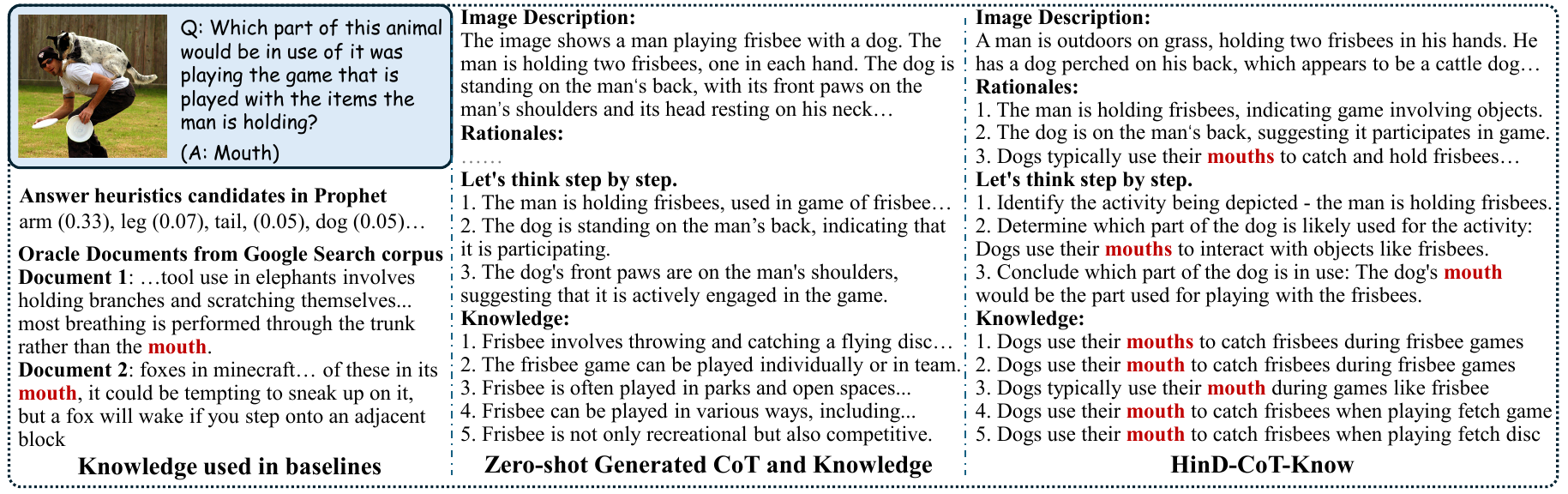}
\vspace{-15pt}
\caption{Case study on OK-VQA test set. The answer heuristics candidates are from the small VQA model MCAN in Prophet \cite{yu2025prophet}. The oracle documents are from Google Search that include the ground-truth answers. } \label{fig:case}
\vspace{-10pt}
\end{figure*}

\subsection{Confidence-Correctness Visualization}
We visualize the joint distribution of the Knowledge Generator's confidence $C$ and correct answer Hits on the OK-VQA test set in Figure \ref{fig:kgen_hits_conf}. 
As shown in Figure \ref{fig:kgen_hits_conf} (a), {Zero-Shot model exhibits over-confidence}: the majority of its generated knowledge has 0 hits (Hits=0), yet its confidence is concentrated at a high level around 0.8. 
Samples with different hits show similar confidence distributions. 
In Figure \ref{fig:kgen_hits_conf} (b), {HinD with HDFT generates knowledge with higher hits, though with an exposed misalignment problem}. The misalignment problem between confidence and correctness show that high-hit knowledge are under-confidence than the 0-hit knowledge. 
In Figure \ref{fig:kgen_hits_conf} (c), {HinD with KEPO effectively corrects the misalignment phenomenon}.
The confidence distribution for high-hit knowledge shifts significantly to the right. 
This indicates that KEPO effectively calibrates the Knowledge Generator, enabling it to be more confident in generating helpful knowledge.

\subsection{Case Study}

Figure \ref{fig:case} shows the case study of HinD and baselines. 

 \textbf{$\bullet$ Comparison with knowledge source in baselines:} SOTA ICL baseline Prophet \cite{yu2025prophet} exploits a smaller VQA model to provide heuristics in example and answer candidates selection, which are limited by the capacity of the heuristic generation model. 
The retrieval-augmented methods with Google Search corpus \cite{luo2021VisDPR} show SOTA PRR@K \cite{lin2023FLMR,long2025reause} in Table \ref{tab: OKVQA}. However, even the oracle documents in the corpus cannot provide relevant knowledge to help reasoning, demonstrating the limitation of the noisy corpus.

 \textbf{$\bullet$ Comparison between Zero-shot and HinD generation:} Zero-shot generated contexts are more diverse, while HinD provides more precise reasoning directions. 
Moreover, the sampled knowledge pieces are more focused on the direct proofs to address the question, rather than irrelevant but image-concentrated content. 
Improving the knowledge generation diversity while maintaining accuracy would be promising for future studies. 

\subsection{Detailed Analyses}

 \textbf{$\bullet$ Cross-MLLM Generalization: }
Table \ref{tab:backbone} shows the cross-scale (Qwen2.5-VL-3B) and cross-series (LLaVA-NeXT-8B and InternVL3-8B) MLLM generalization performance, where the results are from MLLMs distilled from Hindsight-Zero based Qwen2.5-VL-7B.
LLaVA-NeXT exhibits high zero-shot performance, probably due to OK-VQA and A-OKVQA pre-training in LLaVA-1.5 and limits HinD improvements with potential pre-training bias. Qwen2.5-VL and InternVL3 focus on general multimodal capabilities, yielding more gains in learning from Hindsight-Zero, demonstrating cross-backbone generalization.

\begin{table}[!t]
\caption{Cross-Backbone Distillation results.}
\label{tab:backbone}
\begin{center}
\vspace{-15pt}

\resizebox{\linewidth}{!}{
\centering
\small
\begin{tabular}{lccccc}
\toprule
\multirow{2}{*}{MLLM} & Zero-Shot  & \multicolumn{2}{c}{HinD-Know} &\multicolumn{2}{c}{HinD-CoT-Know}  \\
\cmidrule{2-6} 
                 &Score      & PRR@K       & Score  &PRR@K       & Score      \\

       \midrule
   Qwen2.5-VL-7B        & 61.6 & 94.7 & 68.3 & 95.3 & 67.5 \\
Qwen2.5-VL-3B & 60.6 & 94.8 & 64.7 & 95.6 & 65.9 \\
LLaVA-NeXT-8B & \textbf{64.6} & 94.1&64.3 & 95.4 & 65.7 \\
InternVL3-8B         & 61.0 & 95.8 & 68.4  &\textbf{96.7} & \textbf{68.6} \\
\bottomrule
\end{tabular}}
\end{center}
\end{table}

\begin{table}[!t]
\caption{Out-of-distribution experiments on ScienceQA and MME. }
\label{tab: transfer}
\vspace{-15pt}
\begin{center}
\resizebox{\linewidth}{!}{
\centering
\small
\begin{tabular}{c c c}
\toprule
&Zero-shot ScienceQA & A-OKVQA (MC) $\rightarrow$ ScienceQA   \\
       \midrule
Accuracy&71.1 & 83.5 \\
\midrule
&Zero-Shot MME & OK-VQA (DA) $\rightarrow$ MME (Y/N) \\
\midrule
Perception / Cognition &1685.2~/~630.4 & 1683.7~/~716.1\\

\bottomrule
\end{tabular}}
\end{center}
\vspace{-10pt}
\end{table}

 \textbf{$\bullet$ Cross-Dataset Generalization: }
Table \ref{tab: transfer} shows the out of distribution (OOD) generalization of HinD. HinD trained on A-OKVQA is evaluated on ScienceQA \cite{lu2022ScienceQA}. Compared with zero-shot accuracy, the performance is improved by 12.4\%, demonstrating that the reasoning ability of HinD can be transferred to unseen knowledge-intensive tasks without training, confirming robustness.

\textbf{$\bullet$ Cross-Format Generalization: }
Table \ref{tab: transfer} shows the generalization across answering formats. HinD trained on OK-VQA under the open-ended direct-answering setting is evaluated on MME \cite{fu2026mme}, a general VQA benchmark with an unseen yes/no format free from answer string constraints. Compared with zero-shot, the Cognition score is improved by 13.6\% (630.4 to 716.1), while the Perception
score remains stable, demonstrating that HinD strengthens the knowledge reasoning ability without
sacrificing the general perception ability of the MLLM.

 \textbf{$\bullet$ Confusion Matrices and Errors. } 
Figure \ref{fig:okvqa_stats} shows the confusion matrix and statistics of HinD-Know and HinD-CoT-Know on OK-VQA. 
Comparing HinD-Know and HinD-CoT-Know, though HinD-CoT-Know has more hit CoT and Knowledge contexts (95.3\% v.s. 94.7\%), HinD-Know provides better accurate rates. 
The reason could be that CoT brings more relevant contexts, as well as noisy information that interrupts the reasoning. 
As shown in Figure \ref{fig:case}, the Knowledge Generator is more focused on the answer-directed fact illustration, while CoT concentrates more on the reasoning actions. 
Thus, HinD-Know performs better on OK-VQA while HinD-CoT-Know performs better on A-OKVQA which requires more commonsense reasoning \cite{schwenk2022aokvqa}. 

 \textbf{$\bullet$ Error Analysis. }
HinD-Know and HinD-CoT-Know both have large proportions (21.8\% \& 23.3\%) of samples that included relevant contexts but inferred the wrong answers. 
We employ DeepSeek-V4-Flash to categorize these 1,178
context-hit-answer-miss cases into five types, as in Table \ref{tab: errortype}.
\textit{Wrong Direction} and \textit{Noise} dominate,
indicating that the Answer Generator is misled when the hit evidence is buried
among plausible but off-target contexts, while \textit{Lexical} mismatch reflects the annotation granularity of the datasets rather than a reasoning failure. 
\textit{Extraction} error indicates that wrong keywords are extracted.
Notably, the \textit{Coincidental} keyword hits account for 0.0\%, confirming that HinD performs reasoning rather than learning superficial answer patterns. 
Types A, B, and D suggest that improving the noise robustness and summarization ability of the Answer Generator is promising for future studies.

\begin{table}[t]
\caption{Context-hit-answer-miss error statistics, categorized by DeepSeek-V4-Flash.} \label{tab: errortype}
\centering\small
\resizebox{\linewidth}{!}{
\begin{tabular}{ccccc}
\toprule
A. Wrong Direction & B. Noise & C. Lexical & D. Extraction & E. Coincidental \\
\midrule
43.7\% & 33.5\% & 19.8\% & 3.0\% & {0.0\%} \\
\bottomrule
\end{tabular}}
\end{table}

\begin{figure}[!t]
\centering
\includegraphics[width=0.5\textwidth]{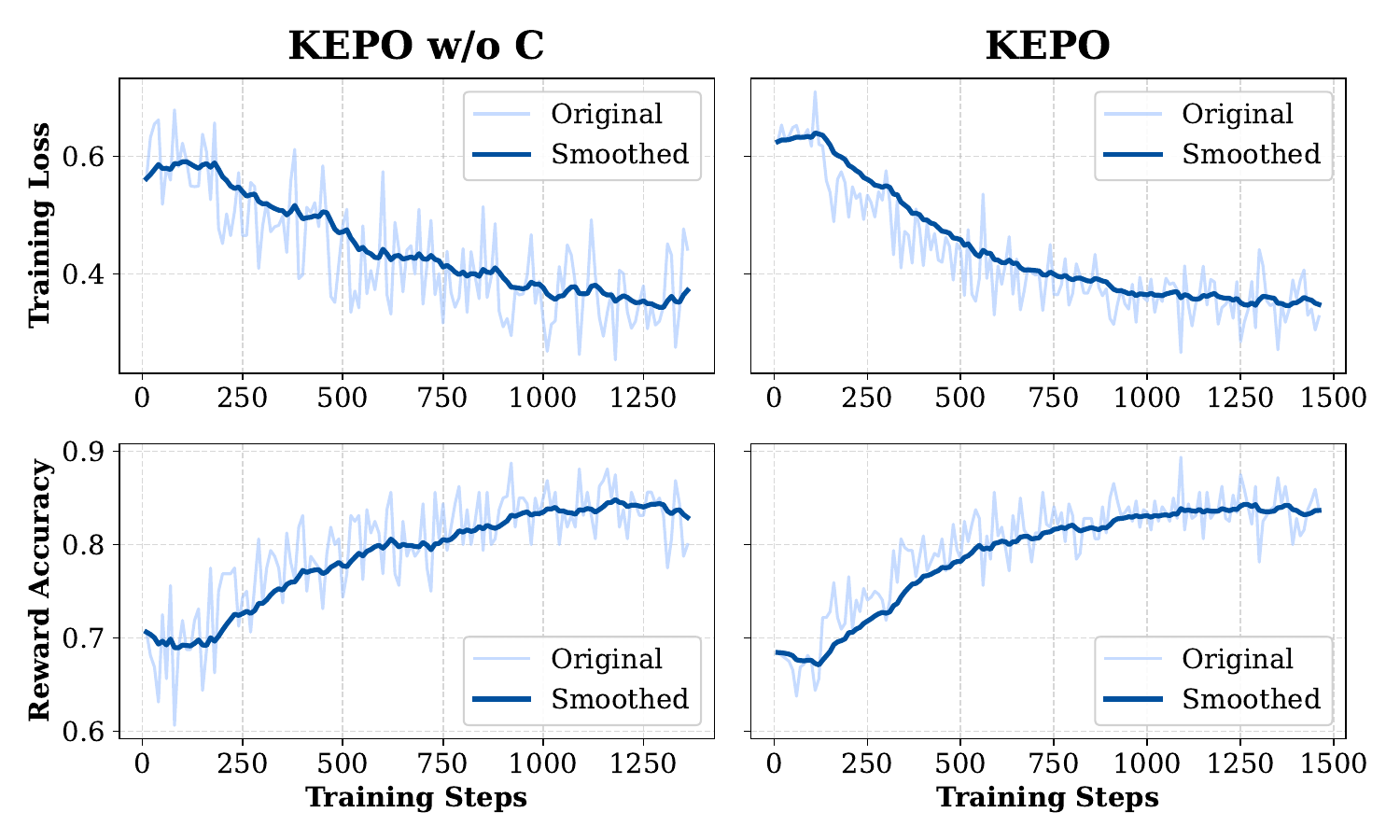}
\caption{Training losses and reward accuracies of KEPO. }  \label{fig:kepo_log}
\end{figure}
\begin{figure}[!t]
\centering
\includegraphics[width=0.4\textwidth]{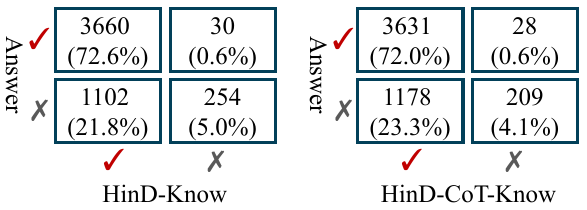}
\caption{Confusion matrix on OK-VQA test set. }  \label{fig:okvqa_stats}
\vspace{-10pt}
\end{figure}

 \textbf{$\bullet$ Confidence for KEPO Optimization:}
As shown in Figure \ref{fig:kepo_log}, KEPO w/ $C$ provides higher initial loss and lower initial reward accuracy. 
It demonstrates that with confidence-aware pair construction, KEPO provides a larger learnable gap for the MLLM, thus yielding more gains after optimization. 
Moreover, the stability of KEPO w/ $C$ is also higher than w/o $C$, demonstrating the superiority of confidence-aware design.

\section{Conclusion}

In this paper, we introduce the Self-Encouraged Hindsight Distillation Reasoning framework, HinD, for KBVQA, aiming at constructing Hindsight Teacher with privileged information to teach Foresight Student.
By treating the ground-truth answer as privileged information available only in training, the Hindsight Teacher completes the intermediate reasoning
process. 
Then, the Foresight Student without access to answers learns from the teacher in two ways: 
Hindsight Distillation Fine-Tuning teaches what to generate from the golden trajectories, while Knowledge
Encouragement Preference Optimization aligns the correctness and confidence.
Experiments demonstrate the superiority of HinD on both VQA Score and knowledge relevance metrics based on 7-B scale MLLMs without commercial model APIs or retrieved knowledge, providing an effective CoT-tuning framework for KBVQA.

\bibliographystyle{IEEEtran}
\bibliography{mybib}

@inproceedings{schwenk2022aokvqa,
  title={A-okvqa: A benchmark for visual question answering using world knowledge},
  author={Schwenk, Dustin and Khandelwal, Apoorv and Clark, Christopher and Marino, Kenneth and Mottaghi, Roozbeh},
  booktitle={European conference on computer vision},
  pages={146--162},
  year={2022},
  organization={Springer}
}

@InProceedings{Fang_2025_CVPR_NoteMR,
    author    = {Fang, Wenlong and Wu, Qiaofeng and Chen, Jing and Xue, Yun},
    title     = {Notes-guided MLLM Reasoning: Enhancing MLLM with Knowledge and Visual Notes for Visual Question Answering},
    booktitle = {Proceedings of the IEEE/CVF Conference on Computer Vision and Pattern Recognition (CVPR)},
    month     = {June},
    year      = {2025},
    pages     = {19597-19607}
}

@inproceedings{yang2025QACap,
  title={Separation of powers: On segregating knowledge from observation in LLM-enabled knowledge-based visual question answering},
  author={Yang, Zhen and Tao, Zhuo and Chen, Qi and Li, Liang and Qi, Yuankai and van den Hengel, Anton and Huang, Qingming},
  booktitle={Proceedings of the Computer Vision and Pattern Recognition Conference},
  pages={24753--24762},
  year={2025}
}

@inproceedings{marino2019okvqa,
  title={Ok-vqa: A visual question answering benchmark requiring external knowledge},
  author={Marino, Kenneth and Rastegari, Mohammad and Farhadi, Ali and Mottaghi, Roozbeh},
  booktitle={Proceedings of the IEEE/cvf conference on computer vision and pattern recognition},
  pages={3195--3204},
  year={2019}
}

@article{rafailov2023DPO,
  title={Direct preference optimization: Your language model is secretly a reward model},
  author={Rafailov, Rafael and Sharma, Archit and Mitchell, Eric and Manning, Christopher D and Ermon, Stefano and Finn, Chelsea},
  journal={Advances in neural information processing systems},
  volume={36},
  pages={53728--53741},
  year={2023}
}

@article{yu2025prophet,
  title={Prophet: Prompting large language models with complementary answer heuristics for knowledge-based visual question answering},
  author={Yu, Zhou and Ouyang, Xuecheng and Shao, Zhenwei and Wang, Meng and Yu, Jun},
  journal={IEEE Transactions on Pattern Analysis and Machine Intelligence},
  year={2025},
  publisher={IEEE}
}

@inproceedings{long2025reause,
  title={Retrieval-Augmented Visual Question Answering via Built-in Autoregressive Search Engines},
  author={Long, Xinwei and Ma, Zhiyuan and Hua, Ermo and Zhang, Kaiyan and Qi, Biqing and Zhou, Bowen},
  booktitle={Proceedings of the AAAI Conference on Artificial Intelligence},
  volume={39},
  number={23},
  pages={24723--24731},
  year={2025}
}

@article{team2024qwen2report,
  title={Qwen2.5-VL Technical Report},
  author={Team, Qwen},
  journal = {arXiv preprint arXiv:2502.13923},
  year    = {2025}
}

@article{lin2022revive,
  title={Revive: Regional visual representation matters in knowledge-based visual question answering},
  author={Lin, Yuanze and Xie, Yujia and Chen, Dongdong and Xu, Yichong and Zhu, Chenguang and Yuan, Lu},
  journal={Advances in neural information processing systems},
  volume={35},
  pages={10560--10571},
  year={2022}
}

@inproceedings{garderes2020conceptbert,
  title={Conceptbert: Concept-aware representation for visual question answering},
  author={Gard{\`e}res, Fran{\c{c}}ois and Ziaeefard, Maryam and Abeloos, Baptiste and Lecue, Freddy},
  booktitle={Findings of the Association for Computational Linguistics: EMNLP 2020},
  pages={489--498},
  year={2020}
}

@inproceedings{marino2021krisp,
  title={Krisp: Integrating implicit and symbolic knowledge for open-domain knowledge-based vqa},
  author={Marino, Kenneth and Chen, Xinlei and Parikh, Devi and Gupta, Abhinav and Rohrbach, Marcus},
  booktitle={Proceedings of the IEEE/CVF conference on computer vision and pattern recognition},
  pages={14111--14121},
  year={2021}
}

@article{luo2021VisDPR,
  title={Weakly-supervised visual-retriever-reader for knowledge-based question answering},
  author={Luo, Man and Zeng, Yankai and Banerjee, Pratyay and Baral, Chitta},
  journal={arXiv preprint arXiv:2109.04014},
  year={2021}
}

@inproceedings{wu2022mavex,
  title={Multi-modal answer validation for knowledge-based vqa},
  author={Wu, Jialin and Lu, Jiasen and Sabharwal, Ashish and Mottaghi, Roozbeh},
  booktitle={Proceedings of the AAAI conference on artificial intelligence},
  volume={36},
  number={3},
  pages={2712--2721},
  year={2022}
}

@inproceedings{li2023blip2,
  title={Blip-2: Bootstrapping language-image pre-training with frozen image encoders and large language models},
  author={Li, Junnan and Li, Dongxu and Savarese, Silvio and Hoi, Steven},
  booktitle={International conference on machine learning},
  pages={19730--19742},
  year={2023},
  organization={PMLR}
}

@article{xu2024llava-cot,
  title={Llava-cot: Let vision language models reason step-by-step},
  author={Xu, Guowei and Jin, Peng and Li, Hao and Song, Yibing and Sun, Lichao and Yuan, Li},
  journal={arXiv preprint arXiv:2411.10440},
  year={2024}
}

@article{hu2022lora,
  title={Lora: Low-rank adaptation of large language models.},
  author={Hu, Edward J and Shen, Yelong and Wallis, Phillip and Allen-Zhu, Zeyuan and Li, Yuanzhi and Wang, Shean and Wang, Lu and Chen, Weizhu and others},
  journal={ICLR},
  volume={1},
  number={2},
  pages={3},
  year={2022}
}

@inproceedings{zheng2024llamafactory,
  title={LlamaFactory: Unified Efficient Fine-Tuning of 100+ Language Models},
  author={Yaowei Zheng and Richong Zhang and Junhao Zhang and Yanhan Ye and Zheyan Luo and Zhangchi Feng and Yongqiang Ma},
  booktitle={Proceedings of the 62nd Annual Meeting of the Association for Computational Linguistics (Volume 3: System Demonstrations)},
  address={Bangkok, Thailand},
  year={2024},
}

@inproceedings{shah2019kvqa,
  title={Kvqa: Knowledge-aware visual question answering},
  author={Shah, Sanket and Mishra, Anand and Yadati, Naganand and Talukdar, Partha Pratim},
  booktitle={Proceedings of the AAAI conference on artificial intelligence},
  volume={33},
  number={01},
  pages={8876--8884},
  year={2019}
}

@article{lu2022ScienceQA,
  title={Learn to explain: Multimodal reasoning via thought chains for science question answering},
  author={Lu, Pan and Mishra, Swaroop and Xia, Tanglin and Qiu, Liang and Chang, Kai-Wei and Zhu, Song-Chun and Tafjord, Oyvind and Clark, Peter and Kalyan, Ashwin},
  journal={Advances in Neural Information Processing Systems},
  volume={35},
  pages={2507--2521},
  year={2022}
}

@article{guo2025deepseekR1,
  title={Deepseek-r1: Incentivizing reasoning capability in llms via reinforcement learning},
  author={Guo, Daya and Yang, Dejian and Zhang, Haowei and Song, Junxiao and Zhang, Ruoyu and Xu, Runxin and Zhu, Qihao and Ma, Shirong and Wang, Peiyi and Bi, Xiao and others},
  journal={arXiv preprint arXiv:2501.12948},
  year={2025}
}

@article{raffel2020T5,
  title={Exploring the limits of transfer learning with a unified text-to-text transformer},
  author={Raffel, Colin and Shazeer, Noam and Roberts, Adam and Lee, Katherine and Narang, Sharan and Matena, Michael and Zhou, Yanqi and Li, Wei and Liu, Peter J},
  journal={Journal of machine learning research},
  volume={21},
  number={140},
  pages={1--67},
  year={2020}
}

@inproceedings{yang2022PICa,
  title={An empirical study of gpt-3 for few-shot knowledge-based vqa},
  author={Yang, Zhengyuan and Gan, Zhe and Wang, Jianfeng and Hu, Xiaowei and Lu, Yumao and Liu, Zicheng and Wang, Lijuan},
  booktitle={Proceedings of the AAAI conference on artificial intelligence},
  volume={36},
  number={3},
  pages={3081--3089},
  year={2022}
}

@inproceedings{speer2017conceptnet,
  title={Conceptnet 5.5: An open multilingual graph of general knowledge},
  author={Speer, Robyn and Chin, Joshua and Havasi, Catherine},
  booktitle={Proceedings of the AAAI conference on artificial intelligence},
  volume={31},
  number={1},
  year={2017}
}

@article{vrandevcic2014wikidata,
  title={Wikidata: a free collaborative knowledgebase},
  author={Vrande{\v{c}}i{\'c}, Denny and Kr{\"o}tzsch, Markus},
  journal={Communications of the ACM},
  volume={57},
  number={10},
  pages={78--85},
  year={2014},
  publisher={ACM New York, NY, USA}
}

@inproceedings{gao2022trig,
  title={Transform-retrieve-generate: Natural language-centric outside-knowledge visual question answering},
  author={Gao, Feng and Ping, Qing and Thattai, Govind and Reganti, Aishwarya and Wu, Ying Nian and Natarajan, Prem},
  booktitle={Proceedings of the IEEE/CVF conference on computer vision and pattern recognition},
  pages={5067--5077},
  year={2022}
}

@inproceedings{gui2022kat,
  title={KAT: A Knowledge Augmented Transformer for Vision-and-Language},
  author={Gui, Liangke and Wang, Borui and Huang, Qiuyuan and Hauptmann, Alexander G and Bisk, Yonatan and Gao, Jianfeng},
  booktitle={Proceedings of the 2022 Conference of the North American Chapter of the Association for Computational Linguistics: Human Language Technologies},
  pages={956--968},
  year={2022}
}

@inproceedings{lin2022RAVQA,
  title={Retrieval Augmented Visual Question Answering with Outside Knowledge},
  author={Lin, Weizhe and Byrne, Bill},
  booktitle={Proceedings of the 2022 Conference on Empirical Methods in Natural Language Processing},
  pages={11238--11254},
  year={2022}
}

@inproceedings{si2023TwO,
  title={Combo of Thinking and Observing for Outside-Knowledge VQA},
  author={Si, Qingyi and Mo, Yuchen and Lin, Zheng and Ji, Huishan and Wang, Weiping},
  booktitle={Proceedings of the 61st Annual Meeting of the Association for Computational Linguistics (Volume 1: Long Papers)},
  pages={10959--10975},
  year={2023}
}

@inproceedings{hu2023promptcap,
  title={Promptcap: Prompt-guided image captioning for vqa with gpt-3},
  author={Hu, Yushi and Hua, Hang and Yang, Zhengyuan and Shi, Weijia and Smith, Noah A and Luo, Jiebo},
  booktitle={Proceedings of the IEEE/CVF International Conference on Computer Vision},
  pages={2963--2975},
  year={2023}
}

@inproceedings{khademi2023mmreasoner,
  title={MM-reasoner: A multi-modal knowledge-aware framework for knowledge-based visual question answering},
  author={Khademi, Mahmoud and Yang, Ziyi and Frujeri, Felipe and Zhu, Chenguang},
  booktitle={Findings of the Association for Computational Linguistics: EMNLP 2023},
  pages={6571--6581},
  year={2023}
}

@inproceedings{xenos2023SimKBVQA,
  title={A Simple Baseline for Knowledge-Based Visual Question Answering},
  author={Xenos, Alexandros and Stafylakis, Themos and Patras, Ioannis and Tzimiropoulos, Georgios},
  booktitle={Proceedings of the 2023 Conference on Empirical Methods in Natural Language Processing},
  pages={14871--14877},
  year={2023}
}

@article{jiang2025corvid,
  title={Corvid: Improving Multimodal Large Language Models Towards Chain-of-Thought Reasoning},
  author={Jiang, Jingjing and Ma, Chao and Song, Xurui and Zhang, Hanwang and Luo, Jun},
  journal={arXiv preprint arXiv:2507.07424},
  year={2025}
}

@inproceedings{dong2024MAIL,
  title={Modality-Aware Integration with Large Language Models for Knowledge-based Visual Question Answering},
  author={Dong, Junnan and Zhang, Qinggang and Zhou, Huachi and Zha, Daochen and Zheng, Pai and Huang, Xiao},
  booktitle={62nd Annual Meeting of the Association for Computational Linguistics, ACL 2024},
  pages={2417--2429},
  year={2024},
  organization={Association for Computational Linguistics (ACL)}
}

@inproceedings{yang2024selfdistillbridge,
  title={Self-Distillation Bridges Distribution Gap in Language Model Fine-Tuning},
  author={Yang, Zhaorui and Pang, Tianyu and Feng, Haozhe and Wang, Han and Chen, Wei and Zhu, Minfeng and Liu, Qian},
  booktitle={Proceedings of the 62nd Annual Meeting of the Association for Computational Linguistics (Volume 1: Long Papers)},
  pages={1028--1043},
  year={2024}
}

@inproceedings{antol2015vqa,
  title={Vqa: Visual question answering},
  author={Antol, Stanislaw and Agrawal, Aishwarya and Lu, Jiasen and Mitchell, Margaret and Batra, Dhruv and Zitnick, C Lawrence and Parikh, Devi},
  booktitle={Proceedings of the IEEE international conference on computer vision},
  pages={2425--2433},
  year={2015}
}

@misc{liu2023llava,
      title={Visual Instruction Tuning}, 
      author={Liu, Haotian and Li, Chunyuan and Wu, Qingyang and Lee, Yong Jae},
      publisher={NeurIPS},
      year={2023},
}

@inproceedings{chen2023infoseek,
  title={Can Pre-trained Vision and Language Models Answer Visual Information-Seeking Questions?},
  author={Chen, Yang and Hu, Hexiang and Luan, Yi and Sun, Haitian and Changpinyo, Soravit and Ritter, Alan and Chang, Ming-Wei},
  booktitle={Proceedings of the 2023 Conference on Empirical Methods in Natural Language Processing},
  pages={14948--14968},
  year={2023}
}

@inproceedings{goyal2017vqav2,
  title={Making the v in vqa matter: Elevating the role of image understanding in visual question answering},
  author={Goyal, Yash and Khot, Tejas and Summers-Stay, Douglas and Batra, Dhruv and Parikh, Devi},
  booktitle={Proceedings of the IEEE conference on computer vision and pattern recognition},
  pages={6904--6913},
  year={2017}
}

@article{brown2020GPT3,
  title={Language models are few-shot learners},
  author={Brown, Tom and Mann, Benjamin and Ryder, Nick and Subbiah, Melanie and Kaplan, Jared D and Dhariwal, Prafulla and Neelakantan, Arvind and Shyam, Pranav and Sastry, Girish and Askell, Amanda and others},
  journal={Advances in neural information processing systems},
  volume={33},
  pages={1877--1901},
  year={2020}
}

@article{hurst2024gpt4o,
  title={Gpt-4o system card},
  author={Hurst, Aaron and Lerer, Adam and Goucher, Adam P and Perelman, Adam and Ramesh, Aditya and Clark, Aidan and Ostrow, AJ and Welihinda, Akila and Hayes, Alan and Radford, Alec and others},
  journal={arXiv preprint arXiv:2410.21276},
  year={2024}
}

@article{touvron2023llama2,
  title={Llama 2: Open foundation and fine-tuned chat models},
  author={Touvron, Hugo and Martin, Louis and Stone, Kevin and Albert, Peter and Almahairi, Amjad and Babaei, Yasmine and Bashlykov, Nikolay and Batra, Soumya and Bhargava, Prajjwal and Bhosale, Shruti and others},
  journal={arXiv preprint arXiv:2307.09288},
  year={2023}
}

@article{achiam2023gpt4,
  title={Gpt-4 technical report},
  author={Achiam, Josh and Adler, Steven and Agarwal, Sandhini and Ahmad, Lama and Akkaya, Ilge and Aleman, Florencia Leoni and Almeida, Diogo and Altenschmidt, Janko and Altman, Sam and Anadkat, Shyamal and others},
  journal={arXiv preprint arXiv:2303.08774},
  year={2023}
}

@article{lin2023FLMR,
  title={Fine-grained late-interaction multi-modal retrieval for retrieval augmented visual question answering},
  author={Lin, Weizhe and Chen, Jinghong and Mei, Jingbiao and Coca, Alexandru and Byrne, Bill},
  journal={Advances in Neural Information Processing Systems},
  volume={36},
  pages={22820--22840},
  year={2023}
}

@inproceedings{hao2024boter,
  title={Self-Bootstrapped Visual-Language Model for Knowledge Selection and Question Answering},
  author={Hao, Dongze and Wang, Qunbo and Guo, Longteng and Jiang, Jie and Liu, Jing},
  booktitle={Proceedings of the 2024 Conference on Empirical Methods in Natural Language Processing},
  pages={1857--1868},
  year={2024}
}

@inproceedings{jian2024LLMRA,
  title={Large language models know what is key visual entity: An LLM-assisted multimodal retrieval for VQA},
  author={Jian, Pu and Yu, Donglei and Zhang, Jiajun},
  booktitle={Proceedings of the 2024 Conference on Empirical Methods in Natural Language Processing},
  pages={10939--10956},
  year={2024}
}

@inproceedings{driess2023palme,
  title={PaLM-E: an embodied multimodal language model},
  author={Driess, Danny and Xia, Fei and Sajjadi, Mehdi SM and Lynch, Corey and Chowdhery, Aakanksha and Ichter, Brian and Wahid, Ayzaan and Tompson, Jonathan and Vuong, Quan and Yu, Tianhe and others},
  booktitle={Proceedings of the 40th International Conference on Machine Learning},
  pages={8469--8488},
  year={2023}
}

@article{dai2023instructblip,
  title={Instructblip: Towards general-purpose vision-language models with instruction tuning},
  author={Dai, Wenliang and Li, Junnan and Li, Dongxu and Tiong, Anthony and Zhao, Junqi and Wang, Weisheng and Li, Boyang and Fung, Pascale N and Hoi, Steven},
  journal={Advances in neural information processing systems},
  volume={36},
  pages={49250--49267},
  year={2023}
}

@article{zhu2023minigpt4,
  title={Minigpt-4: Enhancing vision-language understanding with advanced large language models},
  author={Zhu, Deyao and Chen, Jun and Shen, Xiaoqian and Li, Xiang and Elhoseiny, Mohamed},
  journal={arXiv preprint arXiv:2304.10592},
  year={2023}
}

@inproceedings{hu2023reveal,
  title={Reveal: Retrieval-augmented visual-language pre-training with multi-source multimodal knowledge memory},
  author={Hu, Ziniu and Iscen, Ahmet and Sun, Chen and Wang, Zirui and Chang, Kai-Wei and Sun, Yizhou and Schmid, Cordelia and Ross, David A and Fathi, Alireza},
  booktitle={Proceedings of the IEEE/CVF conference on computer vision and pattern recognition},
  pages={23369--23379},
  year={2023}
}

@article{wei2022CoT,
  title={Chain-of-thought prompting elicits reasoning in large language models},
  author={Wei, Jason and Wang, Xuezhi and Schuurmans, Dale and Bosma, Maarten and Xia, Fei and Chi, Ed and Le, Quoc V and Zhou, Denny and others},
  journal={Advances in neural information processing systems},
  volume={35},
  pages={24824--24837},
  year={2022}
}

@article{wang2025multimodalCoTsurvey,
  title={Multimodal chain-of-thought reasoning: A comprehensive survey},
  author={Wang, Yaoting and Wu, Shengqiong and Zhang, Yuecheng and Yan, Shuicheng and Liu, Ziwei and Luo, Jiebo and Fei, Hao},
  journal={arXiv preprint arXiv:2503.12605},
  year={2025}
}

@article{zhang2024TMLR-MM-CoT,
  title={Multimodal Chain-of-Thought Reasoning in Language Models},
  author={Zhang, Zhuosheng and Zhang, Aston and Li, Mu and Zhao, Hai and Karypis, George and Smola, Alex},
  journal={Transactions on Machine Learning Research},
  volume={2024},
  year={2024},
  publisher={Transactions on Machine Learning Research}
}

@article{yao2024mulberry,
  title={Mulberry: Empowering mllm with o1-like reasoning and reflection via collective monte carlo tree search},
  author={Yao, Huanjin and Huang, Jiaxing and Wu, Wenhao and Zhang, Jingyi and Wang, Yibo and Liu, Shunyu and Wang, Yingjie and Song, Yuxin and Feng, Haocheng and Shen, Li and others},
  journal={arXiv preprint arXiv:2412.18319},
  year={2024}
}

@inproceedings{chen2024visualCoT-kbvqa,
  title={Visual chain-of-thought prompting for knowledge-based visual reasoning},
  author={Chen, Zhenfang and Zhou, Qinhong and Shen, Yikang and Hong, Yining and Sun, Zhiqing and Gutfreund, Dan and Gan, Chuang},
  booktitle={Proceedings of the AAAI Conference on Artificial Intelligence},
  volume={38},
  number={2},
  pages={1254--1262},
  year={2024}
}

@inproceedings{wang2022selfconsistency,
  title={Self-Consistency Improves Chain of Thought Reasoning in Language Models},
  author={Wang, Xuezhi and Wei, Jason and Schuurmans, Dale and Le, Quoc V and Chi, Ed H and Narang, Sharan and Chowdhery, Aakanksha and Zhou, Denny},
  year={2022},
  booktitle={The Eleventh International Conference on Learning Representations}
}

@inproceedings{sun2025LSMS25MM,
  title={Large-Small Model Synergy with Multimodal Fine-Grained Heuristics for Knowledge-Based Visual Question Answering},
  author={Sun, Zhongfan and Guo, Kan and Hu, Yongli and Tian, Daxin and Gao, Qingqing and Wang, Jiapu and Gao, Junbin and Sun, Yanfeng and Yin, Baocai},
  booktitle={Proceedings of the 33rd ACM International Conference on Multimedia},
  pages={935--944},
  year={2025}
}

@inproceedings{wang2024SKP24ACL,
  title={Soft knowledge prompt: Help external knowledge become a better teacher to instruct llm in knowledge-based vqa},
  author={Wang, Qunbo and Ji, Ruyi and Peng, Tianhao and Wu, Wenjun and Li, Zechao and Liu, Jing},
  booktitle={Proceedings of the 62nd Annual Meeting of the Association for Computational Linguistics (Volume 1: Long Papers)},
  pages={6132--6143},
  year={2024}
}

@article{DBLP:journals/tmm/VisualCommonsenseCoT,
  author       = {Xinyu Li and
                  Jing Zhao and
                  Tongquan Wei and
                  Shiliang Sun},
  title        = {Visual Context and Commonsense-Guided Causal Chain-of-Thoughts for
                  Visual Commonsense Reasoning},
  journal      = {{IEEE} Trans. Multim.},
  volume       = {28},
  pages        = {2719--2730},
  year         = {2026},
  doi          = {10.1109/TMM.2026.3651070},
}

@article{DBLP:journals/tmm/FacialCoT,
  author       = {Xing Lan and
                  Jian Xue and
                  Ji Qi and
                  Dongmei Jiang and
                  Ke Lu and
                  Tat{-}Seng Chua},
  title        = {ExpLLM: Towards Chain of Thought for Facial Expression Recognition},
  journal      = {{IEEE} Trans. Multim.},
  volume       = {27},
  pages        = {3069--3081},
  year         = {2025},
  doi          = {10.1109/TMM.2025.3557704},

}

@article{DBLP:journals/tmm/KBVQA_Prompting,
  author       = {Lei Zhu and
                  Mengxi Ying and
                  Chengyuan Zhang and
                  Deyin Liu and
                  Lin Yuanbo Wu and
                  Shichao Zhang and
                  Xuelong Li},
  title        = {Multi-Modal Refined Prompting for Advancing Knowledge-Based Visual
                  Question Answering},
  journal      = {{IEEE} Trans. Multim.},
  volume       = {28},
  pages        = {3444--3457},
  year         = {2026},
  doi          = {10.1109/TMM.2026.3660119},
}

@article{DBLP:journals/tmm/BiasKBVQA,
  author       = {Jiachen Lu and
                  Min Jiang and
                  Jun Kong and
                  Danfeng Zhuang and
                  Ming Lu},
  title        = {Mitigating Inherent Bias of Answer Heuristic Based Frameworks in Knowledge-Based
                  Visual Question Answering},
  journal      = {{IEEE} Trans. Multim.},
  volume       = {28},
  pages        = {1744--1755},
  year         = {2026},
  doi          = {10.1109/TMM.2025.3645641},

}

@article{DBLP:journals/tmm/FactVQA,
  author       = {Sen Wu and
                  Guoshuai Zhao and
                  Xueming Qian},
  title        = {Resolving Zero-Shot and Fact-Based Visual Question Answering via Enhanced
                  Fact Retrieval},
  journal      = {{IEEE} Trans. Multim.},
  volume       = {26},
  pages        = {1790--1800},
  year         = {2024},
  doi          = {10.1109/TMM.2023.3289729},

}

@article{DBLP:journals/tmm/MLLM26,
  author       = {Pengpeng Qiang and
                  Hongye Tan and
                  Hu Zhang and
                  Xiaoli Li and
                  Ru Li and
                  Jiye Liang},
  title        = {Mitigating Hallucinations in Large Vision-Language Models via Visual-Enhanced
                  Contrastive Decoding},
  journal      = {{IEEE} Trans. Multim.},
  volume       = {28},
  pages        = {3242--3255},
  year         = {2026},
  doi          = {10.1109/TMM.2026.3651099},

}

@article{DBLP:journals/tmm/LuZS26KnowledgeDialog,
  author       = {Chenyu Lu and
                  Jing Zhao and
                  Shiliang Sun},
  title        = {{DVD:} {A} Debiased Visual Dialog Model via Disentangling Knowledge
                  Features},
  journal      = {{IEEE} Trans. Multim.},
  volume       = {28},
  pages        = {3639--3651},
  year         = {2026},
  doi          = {10.1109/TMM.2026.3654405},

}

@article{DBLP:journals/tmm/WangLJWLLCL26KnowledgeVidelAction,
  author       = {Hao Wang and
                  Fang Liu and
                  Licheng Jiao and
                  Jiahao Wang and
                  Shuo Li and
                  Lingling Li and
                  Puhua Chen and
                  Xu Liu},
  title        = {Learning to Prompt With Refining Text Knowledge for Zero-Shot Video
                  Action Recognition},
  journal      = {{IEEE} Trans. Multim.},
  volume       = {28},
  pages        = {4638--4651},
  year         = {2026},
  doi          = {10.1109/TMM.2026.3660143},
}

@article{DBLP:journals/tmm/OuyangTCZXXW26KnowledgeRecommendation,
  author       = {Kai Ouyang and
                  Chen Tang and
                  Zenghao Chai and
                  Wenhao Zheng and
                  Xiangjin Xie and
                  Xuanji Xiao and
                  Zhi Wang},
  title        = {Breaking the Curse of Knowledge: Towards Effective Multimodal Recommendation
                  Using Knowledge Soft Integration},
  journal      = {{IEEE} Trans. Multim.},
  volume       = {28},
  pages        = {4525--4534},
  year         = {2026},
  doi          = {10.1109/TMM.2026.3654432},

}

@article{vapnik2015learningusingprivilegedinformation,
  title={Learning using privileged information: similarity control and knowledge transfer},
  author={Vapnik, Vladimir and Izmailov, Rauf},
  journal={The Journal of Machine Learning Research},
  volume={16},
  number={1},
  pages={2023--2049},
  year={2015},
  publisher={JMLR. org}
}

@article{fu2026mme,
  title={Mme: A comprehensive evaluation benchmark for multimodal large language models},
  author={Fu, Chaoyou and Chen, Peixian and Shen, Yunhang and Qin, Yulei and Zhang, Mengdan and Lin, Xu and Yang, Jinrui and Zheng, Xiawu and Li, Ke and Sun, Xing and others},
  journal={Advances in Neural Information Processing Systems},
  volume={38},
  year={2025}
}

@inproceedings{zelikman2022star,
  title     = {{STaR}: Bootstrapping Reasoning with Reasoning},
  author    = {Zelikman, Eric and Wu, Yuhuai and Mu, Jesse and Goodman, Noah D.},
  booktitle = {Advances in Neural Information Processing Systems},
  volume    = {35},
  pages     = {15476--15488},
  year      = {2022}
}

@inproceedings{lopezpaz2016unifyingprivileged,
  title     = {Unifying Distillation and Privileged Information},
  author    = {Lopez-Paz, David and Bottou, L{\'e}on and Sch{\"o}lkopf, Bernhard and Vapnik, Vladimir},
  booktitle = {International Conference on Learning Representations},
  year      = {2016}
}

@article{deepseek2026v4,
  title   = {{DeepSeek-V4}: Towards Highly Efficient Million-Token Context Intelligence},
  author  = {{DeepSeek-AI} and others},
  journal = {arXiv preprint arXiv:2606.19348},
  year    = {2026},
  doi     = {10.48550/arXiv.2606.19348}
}
\vfill
\newpage
\appendices

\section{Hindsight-Zero Data Construction Details}

We present the statistics of OK-VQA and A-OKVQA, and our generated Hindsight-Zero dataset in Table \ref{tab: dataset}.

For the CoT and Knowledge HDFT stage, we filter the CoT and knowledge piece with ground truths so that CoT and Knowledge HDFT is based on only reasoning process mentioning answers, to provide reasoning heuristic, thus the training samples are lower than the training set.
For the KEPO stage, we construct 0-2 chosen-rejected pairs for each sample.
For Hindsight-Zero samples that only includes accurate knowledge, we construct negative knowledge from random piece generated from other samples. 
Hindsight-Zero samples with only unhelpful knowledge are ommitted. 
We sample 2 pairs for samples has various accurate/wrong situations to increase the diversity of KEPO to avoid overfitting. 
This results in a larger dataset for preference optimization. 
The Answer Generator is trained on the full training sets with also the CoT or knowledge that do not mention answers.
It provides reasoning robustness since it mimics the real inference scene when CoT and knowledge could be less accurate. 

Table \ref{tab:case_hindsight_zero} provides a concrete example of the generated Hindsight-Zero data for CoT HDFT, Knowledge HDFT, and the constructed KEPO pair, illustrating the ``Chosen'' and ``Rejected'' samples.

\section{Hyperparameter Analysis}

\noindent \textbf{$\bullet$ Effect of Knowledge Shots: }
Figure \ref{fig:params_plt} (a) shows the effect of the count of knowledge pieces, $|\hat{\mathcal{K}}|$, in HinD-Know. 
The PRR@K increases with more knowledge pieces, showing that by sampling more knowledge pieces from HinD-Know, it is more likely to involve the ground-truth answers, which would be more helpful to solve the question. 
The best and second-best VQA scores are with $|\hat{\mathcal{K}}|$=1 and $|\hat{\mathcal{K}}|$=8. 
With $|\hat{\mathcal{K}}|$=1, the knowledge is more focused and contexts are simpler for the Answer Generator to understand. 
With more knowledge pieces, the relevant knowledge increases, as well as the noise, which increases the difficulty for the Answer Generator to reason. 
Thus, the VQA score decreases after $|\hat{\mathcal{K}}|$=8. 
In summary, it is important to balance relevance and involved noise of the generated knowledge for better reasoning ability.

\noindent \textbf{$\bullet$ Effect of Hyperparameters: }
We analyze the effect of the KEPO loss coefficient $\beta$ and the sampling temperature in Figure \ref{fig:params_plt} (b). For $\beta$, we observe that a smaller value (0.1) achieves the highest PRR@K. Increasing $\beta$ to 1.0 causes a significant drop in recall (89.6), suggesting that constraining the KEPO policy $\pi_{\phi}^{KEPO}$ to stay close to the reference policy $\pi_{ref}$ is crucial for maintaining knowledge quality. 
For sampling temperature, a very low value (0.3) limits diversity and results in lower recall (91.5). Temperatures of 0.7 (94.3) and 1.0 (94.7) yield the best performance, indicating that moderate stochasticity is beneficial for knowledge generation. 
Moreover, temperature=1.0 provides the best PRR@K results, indicating that improving knowledge diversity is beneficial for KBVQA reasoning. 
\begin{table}[!t]
\caption{Dataset Statistics.}
\label{tab: dataset}
\begin{center}
\resizebox{0.8\linewidth}{!}{
\centering
\small
\begin{tabular}{lrr}
\toprule
dataset & OK-VQA & A-OKVQA  \\
       \midrule
train set        & 9,009   & 17,055   \\
test/valid set   & 5,046 & 1,145 \\
\midrule
\multicolumn{3}{c}{\textbf{\textit{Hindsight-Zero}}} \\
CoT HDFT         & 7,694   & 15,746   \\
Knowledge HDFT   & 8,833   & 16,730   \\
Knowledge KEPO     & 15,636  & 30,203   \\
Answer Generator & 9,009   & 17,055  \\
\bottomrule
\end{tabular}}

\end{center}
\end{table}

\begin{figure}[!t]
\centering
\includegraphics[width=0.5\textwidth]{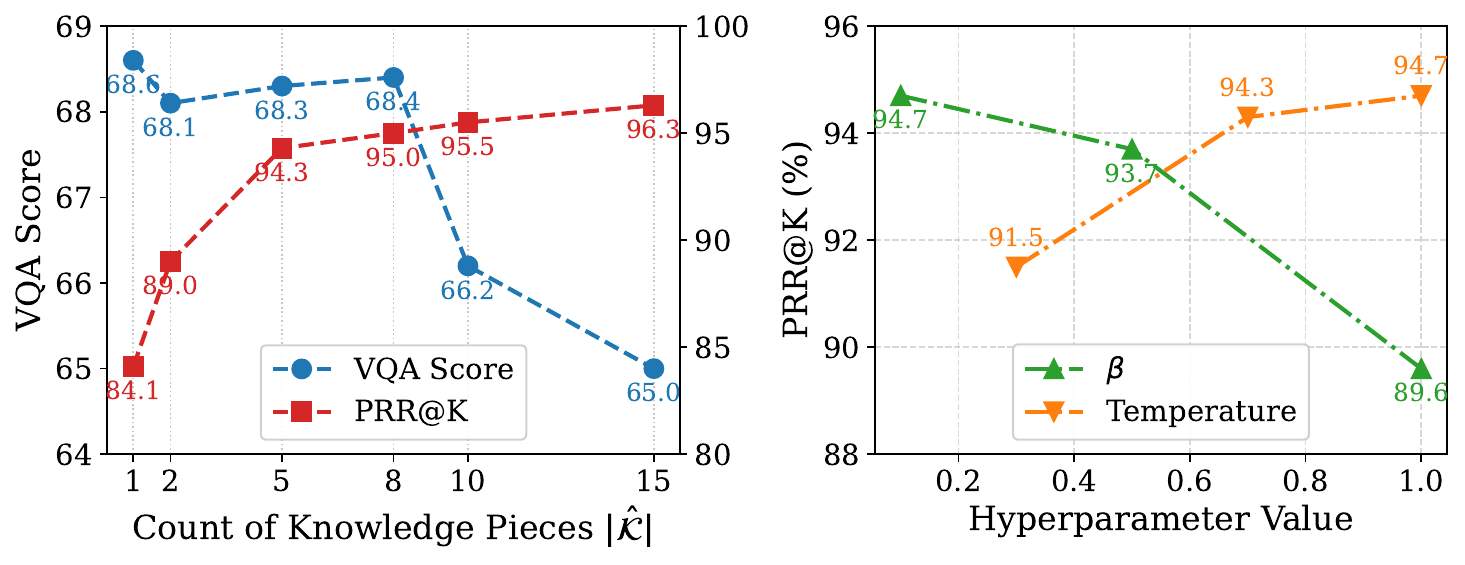}
\caption{Parameter Analysis. (a) Effect of sampled knowledge piece count. (b) Hyperparameter KEPO $\beta$ and sampling temperature.}  \label{fig:params_plt}
\end{figure}

\begin{table}[t]
\caption{Qwen3 Zero-shot VQA Score on OK-VQA. }
\label{tab: qwen3}
\begin{center}
\resizebox{0.6\linewidth}{!}{
\centering
\small
\begin{tabular}{c c}
\toprule
Model & Score \\
       \midrule
Qwen2.5-VL-7B & 61.6 \\
Qwen3-VL-8B-Instruct & 60.6 \\
Qwen3-VL-8B-Thinking & 56.6 \\
\bottomrule
\end{tabular}}
\end{center}
\end{table}

\begin{table}[t]
\caption{HinD with generative retrieval from Google Search corpus on OK-VQA. }
\label{tab: gen_retrieval}
\begin{center}
\resizebox{0.6\linewidth}{!}{
\centering
\small
\begin{tabular}{l c}
\toprule
Model & PRR@K \\
       \midrule
HinD-Know & 94.7 \\
~~~~w/ generative retrieval & 74.9 \\
\bottomrule
\end{tabular}}

\end{center}
\end{table}

\section{Discussions about Thinking model}

We investigate whether a pre-trained Thinking model can solve the KBVQA task in a zero-shot manner. As shown in Table \ref{tab: qwen3}, the latest Qwen3-VL-8B-Thinking model performs significantly worse (56.6) on OK-VQA than its corresponding Instruct model (60.6) and the Qwen2.5-VL-7B base model (61.6). The case study in Table \ref{tab: case_qwen3thinking} illustrates this failure: the Thinking model gets confused by the question's phrasing (``running around the bases on a single hit''), engages in a flawed reasoning process, and incorrectly concludes the answer by rephrasing the question: ``single''. In contrast, our HinD-CoT-Know framework, guided by hindsight-distilled knowledge and CoT, correctly identifies the answer as ``home run''. 
This demonstrates that generic thinking capabilities from the thinking model are unstructured, while the structured sequential thinking steps and discrete knowledge recall from HinD elicited by HDFT and KEPO would reinforce the model to generate targeted, well-organized reasoning processes for better reasoning performance.

\noindent
\begin{table*}[b]
\caption{Case for HinD-Know with generative retrieval from ReAuSE.}
\label{tab: case_gen_retrieval}

    \centering
    \begin{tabularx}{\textwidth}{ L{0.3\textwidth} Y }
\toprule

\multirow{2}{0.3\textwidth}{
    \RaggedRight 
    \textbf{Question}: What could this gentleman be carrying in that red bag? \par \vspace{2mm} 
    \includegraphics[width=\linewidth]{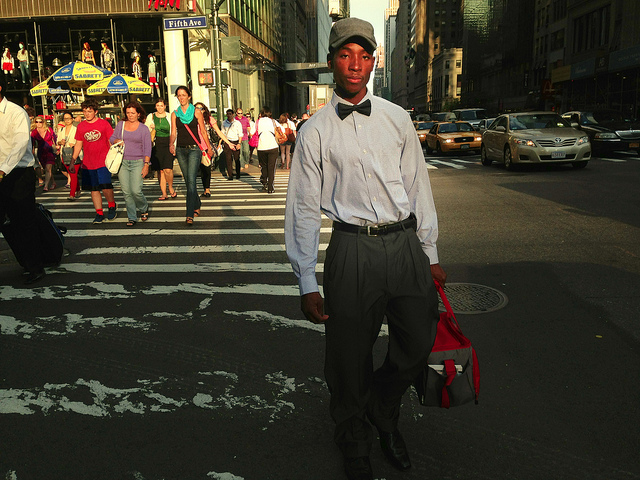} 
    \par \vspace{2mm}
    \textbf{Answer}: clothes/lunch/food/shoes
}



&

\textbf{\textit{HinD-Know}}: 

Knowledge 1: red bag could contain personal items like clothes or electronics but more likely contains lunchbox or other food items for work or school.

Knowledge 2: laptops are commonly carried in bags for work or study but they don't typically match color scheme of red bag well unless it's specialized case like laptop sleeve which might be black or gray with red accents but not entirely red bag itself is unlikely to hold laptop.

Knowledge 3: laptop bags are commonly used for carrying computers which are often used for work or school purposes.

Knowledge 4: laptops and electronics often come in backpacks but they don't typically match red color of bag as well as books might do.

Knowledge 5: luggage is commonly carried for travel purposes and could include clothes or personal items for trip.

\par\vspace{2mm}

\\ \addlinespace\cline{2-2}

&
\vspace{1pt}
\textbf{\textit{HinD-Know w/ generative retrieval}}: 
\par\vspace{2mm}

Document 1: laptops can be warm if it's been operating, and cats love warmth.  this is evidenced by them laying out in sunlight even during the middle of summer.

Document 2: processors for laptops, desktops, servers, and ai intel home toggle navigation sign in sign in username your username is missing password your password is missing by signing in, you agree to our terms of service.  intel® processors bring you world-class performance for business and personal use.  find a wide range of processors by device type—laptops, desktops, workstations, and servers.

Document 3: laptops can be warm if it's been operating, and cats love warmth.  this is evidenced by them laying out in sunlight even during the middle of summer.

Document 4: luggage is constructed to protect the items during travel, either with a hard shell or a durable soft material. luggage often has internal subdivisions or sections to aid in securing items.

Document 5: laptop computers and tablets do not have the adjustability of a desktop computer when adjusting keyboard, mouse and monitor.  “ergonomic” or “natural” keyboards: there are a variety of keyboard types available for use.  as with chairs, desks or other office furniture, sit-stand desks are purchases made at the discretion of the department.

\\ \bottomrule
\end{tabularx}    
\end{table*}

\section{HinD with Google Search Corpus}

To compare our internal knowledge elicitation with external retrieval, we added a constrained decoding procedure following ReAuSE after our Knowledge Generator for a generative retrieval based on Google Search. 
As shown in Table \ref{tab: gen_retrieval}, this approach (w/ generative retrieval) achieves only 74.9 PRR@K, a significant drop from the 94.7 PRR@K of our standard HinD-Know model. 
The case study on Table \ref{tab: case_gen_retrieval} highlights the reason: for a question about items in a red bag, HinD-Know generates relevant, commonsense possibilities like ``lunchbox'' or ``clothes''. In contrast, the external retrieval system returns noisy and irrelevant documents about ``cats'' loving ``laptops'' and ``Intel processors''. 
The reason could be that the MLLM's intuition is laptops, thus MLLM would generatively decode from the corpus for laptop-related passages, which may bring noise and not be helpful. 
This demonstrates that for these KBVQA tasks, the MLLM's internal, distilled knowledge (elicited via HinD) is more concise and relevant than noisy, general-purpose external documents. 

\section{Details of Prompts}

We provide the specific prompt templates used in our framework. 

Listing \ref{lst:prompt_hindsight} details the structured prompt used for generating the Hindsight-Zero training data, which instructs the model to produce an image description, rationales, step-by-step reasoning, and related knowledge pieces given both the question and the ground-truth answer. 
Listing \ref{lst:prompt_cot_generation} and Listing \ref{lst:prompt_kgen} show the prompts used during training and inference for the CoT Generator and the Knowledge Generator, respectively. 
Listing \ref{lst:prompt_ans_gen_cot_know} and Listing \ref{lst:prompt_ans_gen_know} detail the prompts for the final Answer Generator, configured for the HinD-CoT-Know and HinD-Know variants.

\begin{lstlisting}[
    style=promptstyle,
    caption={Prompt used for CoT Generator.},
    label={lst:prompt_cot_generation}
]
Generate an image description based on the question.
Then, provide a rationale to analyze the question.
Next, generate a step-by-step reasoning process to solve the problem. Ensure the steps are logical and concise.

Question: {question}
\end{lstlisting}

\begin{lstlisting}[
    style=promptstyle,
    caption={Prompt used for Knowledge Generator.},
    label={lst:prompt_kgen}
]
Based on the image, generate related knowledge or facts to solve the question. 
Question: {question}
\end{lstlisting}

    \begin{lstlisting}[
    style=promptstyle,
    caption={Prompt used to generate Hindsight-Zero data.},
    label={lst:prompt_hindsight}
]
Generate an image description based on the question.
Then, provide a rationale to analyze the question to give the given answer. 
Next, generate a step-by-step reasoning process to solve the problem towards the given answer. Ensure the steps are logical and concise.
After that, generate 5 related knowledge or facts to solve the question. Ensure the steps are logical and concise.
Finally, provide a concise summary of the final answer with a single word or phrase in the following format: 'The final answer is: xxx'. Do not provide any explanation.

Format your response with the following sections, separated by ###:
### Image Description:
### Rationales:
### Let's think step by step.
### Step 1:
### Step 2:
...
### Knowledge: 
### Knowledge 1:
### Knowledge 2:
...
### The final answer is: 

{question}
{answer}
\end{lstlisting}

\begin{lstlisting}[
    style=promptstyle,
    caption={Prompt used for HinD-CoT-Know Answer Generator.},
    label={lst:prompt_ans_gen_cot_know}
]
Based on the image and reasoning process, respond to this question with a single word or phrase.
Question: {question}

{cot_process}

Knowledge:
{knowledge}

Question: {question}
\end{lstlisting}

\begin{lstlisting}[
    style=promptstyle,
    caption={Prompt used for HinD-Know Answer Generator.},
    label={lst:prompt_ans_gen_know}
]
Based on the image and reasoning process, respond to this question with a single word or phrase.
Question: {question}

Knowledge:
{knowledge}

Question: {question}
\end{lstlisting}


\noindent
\begin{table*}[p]
    \caption{Case for Hindsight-Zero data construction. }
    \label{tab:case_hindsight_zero}
    \centering
\begin{tabularx}{\textwidth}{ L{0.3\textwidth} Y }
\toprule

\multirow{3}{0.3\textwidth}{
    \RaggedRight 
    \textbf{Question}: What is the hairstyle of the blond called? \par \vspace{2mm} 
    \includegraphics[width=\linewidth]{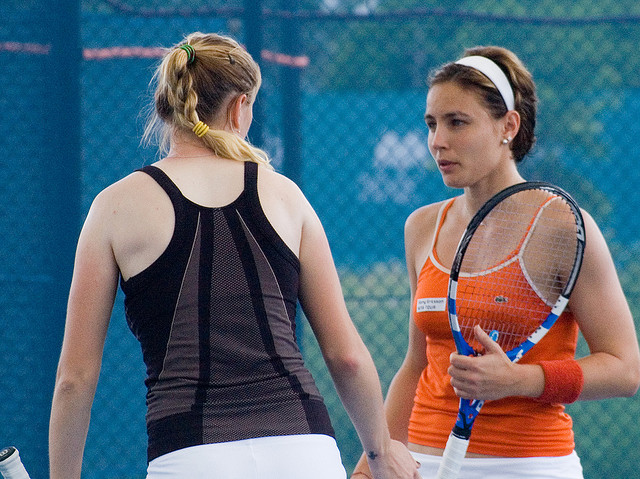} 
    \par \vspace{2mm}
    \textbf{Answer}: ponytail
}

&


\vspace{1pt}
\textbf{\textit{Hindsight-Zero for CoT HDFT}}: 
\par\vspace{2mm}

Image Description:

Two women are standing on a tennis court. One woman has her hair tied back in a ponytail, while the other has her hair down. The woman with the ponytail is wearing a black tank top, and the other is wearing an orange tank top with a white headband.

Rationales:

- The woman with the ponytail has her hair pulled back into a single, thick strand that extends from the back of her head.

- The term \"ponytail\" refers to a hairstyle where hair is gathered at the nape of the neck and secured with a rubber band or similar fastener.

Let's think step by step.

Step 1:
Identify the hairstyle of the woman with the ponytail.

Step 2:
Determine if the hairstyle matches the definition of a ponytail.

Step 3:Conclude whether the hairstyle is correctly identified as a ponytail.

\\ \addlinespace \cline{2-2}

& 
\vspace{1pt}
\textbf{\textit{Hindsight-Zero for Knowledge HDFT}}: 
\par\vspace{2mm}

ponytails are common hairstyle for tennis players due to their practicality and ease of movement during game

\\ \addlinespace\cline{2-2}

& 
\vspace{1pt}
\textbf{\textit{Hindsight-Zero for Knowledge KEPO}}: 
\par\vspace{2mm}

\textit{Chosen}: ponytails are common hairstyle for tennis players due to their practicality and ease of movement during game

\textit{Rejected}: pigtails are common hairstyle for tennis players due to their practicality and ease of movement during game

\\ \bottomrule
\end{tabularx}

\end{table*}

\noindent
\begin{table*}[p]
\caption{Case for HinD-CoT-Know and latest small thinking model Qwen3-8B-Thinking. }
\label{tab: case_qwen3thinking}
    \centering
\begin{tabularx}{\textwidth}{ L{0.3\textwidth} Y }
\toprule

\multirow{2}{0.3\textwidth}{
    \RaggedRight 
    \textbf{Question}: What do they call running around the bases on a single hit? \par \vspace{2mm} 
    \includegraphics[width=\linewidth]{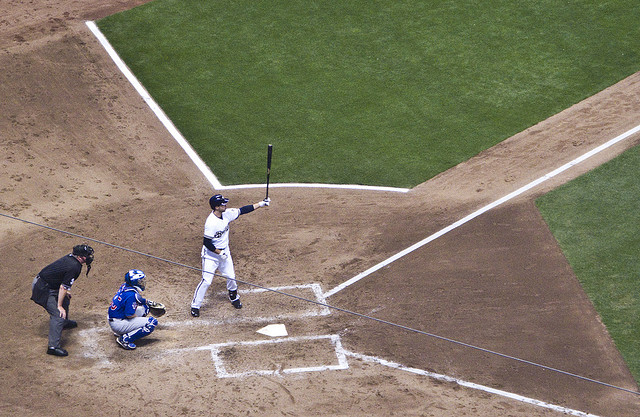} 
    \par \vspace{2mm}
    \textbf{Answer}: home run
}



&

\textbf{\textit{Qwen3-8B-Thinking}}: 
\par\vspace{2mm}
$<think>$
Got it, let's see. The question is about what they call running around the bases on a single hit. In baseball, when a batter hits the ball and gets a single, they run around first base. The term for that is "single," but wait, the question is asking for the name of the action. Wait, no, the action of running around the bases after a single is just part of the play, but the term for the hit itself is "single." Wait, but the question says "running around the bases on a single hit." So when you hit a single, you run around first base. The term for that specific play is a "single," but maybe the answer is "single"? Wait, no, let me think. Wait, in baseball, a single is a hit where the batter reaches first base safely without the ball being fielded and thrown to first. So the act of running around the bases on a single would be... well, the hit is a single, and then you run to first. But the term for that type of hit is "single." So the answer should be "single."
$</think>$

single
\\ \addlinespace\cline{2-2}

&
\vspace{1pt}
\textbf{\textit{HinD-CoT-Know}}: 
\par\vspace{2mm}

\textbf{Image Description}: ...

% ...

\textbf{Rationales}: ...

% ... 




\textbf{Let's think step by step.}

Step 1:
Identify the elements in the image: batter, catcher, umpire, and home plate.

Step 2:
Determine the context: a baseball game in progress.

Step 3:
Consider the question: what do they call running around the bases on a single hit?

Step 4:

Recall the terminology used in baseball: "running around the bases" refers to scoring after hitting a base hit.

\textbf{Knowledge}:

Knowledge 1: baseball term for running around all 4 bases after hitting single is home run but it's not what they call it after hitting single hit is called base hit or single hit is called base hit or single when player hits ball into fair territory and reaches first base safely without being put out

Knowledge 2: baseball term for hitting ball and running around all 4 bases is home run not single hit around bases which is called base hit or base hit with extra bases is triple or home run

Knowledge 3: in baseball term for running around all 4 bases after hitting ball is home run not single hit

Knowledge 4: baseball term for running around all 4 bases after hitting single is home run not single hit

Knowledge 5: baseball term for running around all 4 bases after hitting single is home run not round trip or base hit

\textbf{Answer Generator}: ['home run', 'home run', 'home run', 'base hit', 'base hit'] $\rightarrow$ home run

\\ \bottomrule
\end{tabularx}

\end{table*}

\end{document}